%% file: main.tex
\pdfoutput=1
\documentclass{article} 
\usepackage[T1]{fontenc}
\usepackage[preprint]{colm2025_conference} 

\usepackage{microtype}
\usepackage{hyperref}
\usepackage{url}
\usepackage{booktabs}

\usepackage{lineno}

\definecolor{darkblue}{rgb}{0, 0, 0.5}
\hypersetup{colorlinks=true, citecolor=darkblue, linkcolor=darkblue, urlcolor=darkblue}

\usepackage{longtable}
\usepackage{subcaption}
\newcommand{\methodname}{\texttt{MedScore}}
\newcommand{\datasetname}{\texttt{AskDocsAI}}
\newcommand{\taxonomy}{\texttt{\methodname Taxonomy}}
\usepackage[]{todonotes}
\setuptodonotes{inline}
\usepackage{enumerate}
\usepackage{enumitem}
\usepackage{cleveref}
\usepackage{amsfonts}
\usepackage{multirow}
\usepackage{array}

\title{\methodname: Generalizable Factuality Evaluation of Free-Form Medical Answers by Domain-adapted Claim Decomposition and Verification}

\author{Heyuan Huang, Alexandra DeLucia, Vijay Murari Tiyyala \& Mark Dredze\thanks{This research was, in part, funded by the Advanced Research Projects Agency for Health (ARPA-H). The views and conclusions contained in this document are those of the authors and should not be interpreted as representing the official policies, either expressed or implied, of the United States Government.} \\
Center for Language and Speech Processing\\
Johns Hopkins University\\
Baltimore, MD 21218, USA \\
\texttt{\{hhuan134, aadelucia, vtiyyal1, mdredze\}@jhu.edu} 
}

\begin{document}

\ifcolmsubmission
\linenumbers
\fi

\maketitle

\begin{abstract}
While Large Language Models (LLMs) can generate fluent and convincing responses, they are not necessarily correct. This is especially apparent in the popular decompose-then-verify factuality evaluation pipeline, where LLMs evaluate generations by decomposing the generations into individual, valid claims. 
Factuality evaluation is especially important for medical answers, since incorrect medical information could seriously harm the patient. However, existing factuality systems are a poor match for the medical domain, as they are typically only evaluated on objective, entity-centric, formulaic texts such as biographies and historical topics. This differs from condition-dependent, conversational, hypothetical, sentence-structure diverse, and subjective medical answers, which makes decomposition into valid facts challenging. We propose \methodname, a new pipeline to decompose medical answers into condition-aware valid facts and verify against in-domain corpora. Our method extracts up to three times more valid facts than existing methods, reducing hallucination and vague references, and retaining condition-dependency in facts. The resulting factuality score substantially varies by decomposition method, verification corpus, and used backbone LLM, highlighting the importance of customizing each step for reliable factuality evaluation by using our generalizable and modularized pipeline for domain adaptation.\footnote{The dataset and code are available at \url{https://github.com/Heyuan9/MedScore} }
\end{abstract}

\section{Introduction}
Factuality-based evaluations of generative AI systems ensure that models produce accurate information. 
A common approach to factuality evaluation relies on a \textit{decompose-then-verify} strategy, where individual sentences are first \textit{decomposed} into ``atomic facts'' and then \textit{verified} against a knowledge source \citep{min-etal-2023-factscore}. The final ``factuality'' score is the proportion of claims that can be verified. The popular FActScore framework \citep{min-etal-2023-factscore} exemplifies this approach, which uses Large Language Models (LLMs) to both decompose and verify claims. 
Subsequent work has shown the importance of the decomposition step, as the ability to correctly decompose a generation into atomic facts directly influences the resulting factuality score \citep{wanner-etal-2024-closer}.
Existing frameworks have been evaluated almost exclusively on biographical data, where an LLM-generated biography of a single person is verified against Wikipedia. VeriScore \citep{song-etal-2024-veriscore} expands the generation domains to include fictional stories, Wikipedia, biographies, and an online forum, Reddit.
However, the evaluation of these methods on more complex domains is limited.

LLMs have demonstrated medical ability through tasks such as medical licensing exams \citep{nori2023capabilitiesgpt4medicalchallenge}, answering clinical questions from patient records \citep{pampari-etal-2018-emrqa,PATRICK2012292, Bai:2024ab}, and reasoning about clinical case studies \citep{complex-clinical-gpt4, chen-etal-2025-benchmarking}.
These tasks ask the model to make a decision (e.g., which answer is correct) and are thus scored using traditional accuracy metrics. However, open-ended medical generation tasks require a more sophisticated approach. 
For example, recent work has demonstrated that LLMs can effectively answer patient medical questions \citep{ayers-comparing}. \citet{createTRUST} identified several factors for evaluation of patient-facing systems, including personalization, perceived empathy \citep{machine-empathy}, and factuality. These factors are especially important given the potential harm of incorrect medical information and that LLMs have been known to add irrelevant or sometimes dangerous text \citep{GIORGI2024116058, mei-etal-2023-assert}. 
Decompose-then-verify frameworks may be well-suited to evaluating patient-facing medical Generative AI systems.

However, the focus on biographical text makes current systems a poor match for the medical domain.
Biographical text generations contain
entity-centric, context-free, formulaic sentence structures (e.g., ``Michael Collins was a pilot''), while medical answers are more context-dependent with complex sentence structures. 
Medical answers usually contain subjective-sounding suggestions (``I recommend you take ibuprofen''), hypotheticals (``If the pain gets worse, you should call a doctor''), imperatives (``Start with small steps'') and context-dependent statements (``Ice is good for this situation'') that can be hard to verify in their current form. Existing systems fail to accurately decompose and verify this type of data.

We explore the efficacy and limitations of current factuality evaluation methods for two medical free-form QA datasets: \datasetname, our newly introduced medical QA dataset from Reddit, and PUMA \citep{naik-etal-2024-perspective}, and one non-medical dataset, CaLMQA \citep{arora-etal-2025-calmqa}. We evaluate four existing decomposition and claim filtering methods and show limitations in their approaches (FActScore \citep{min-etal-2023-factscore}, VeriScore \citep{song-etal-2024-veriscore}, Core\citep{jiang_core_2024}, DnDScore\citep{wanner_dndscore_2024}). We present and evaluate \methodname, a new domain-adapted decompose-then-verify pipeline designed for the medical setting that provides the best coverage of condition-aware valid medical claims. Our work shows the importance of customizing data-adapted prompts for claim decompositions before factuality evaluation, and we discuss challenges and directions for medical text evaluation in error analysis in \Cref{sec:results_verify}.

Our contributions are as follows.

\begin{enumerate}[noitemsep]
    \item \taxonomy, a taxonomy for claim decomposition for patient-facing medical text evaluation that highlights existing challenges;
    \item \methodname, a new taxonomy-based 
    factuality evaluation pipeline for free-form medical answers with fewer invalid claims than FActScore and higher coverage than VeriScore, which is generalizable to other domains with minor change;
    \item \datasetname, a high-quality free-response medical Question-Answer (QA) dataset with real user questions and LLM-augmented doctor answers from Reddit's r/AskDocs.
\end{enumerate}

\begin{figure*}[t]
    \centering
    \includegraphics[width=\textwidth]{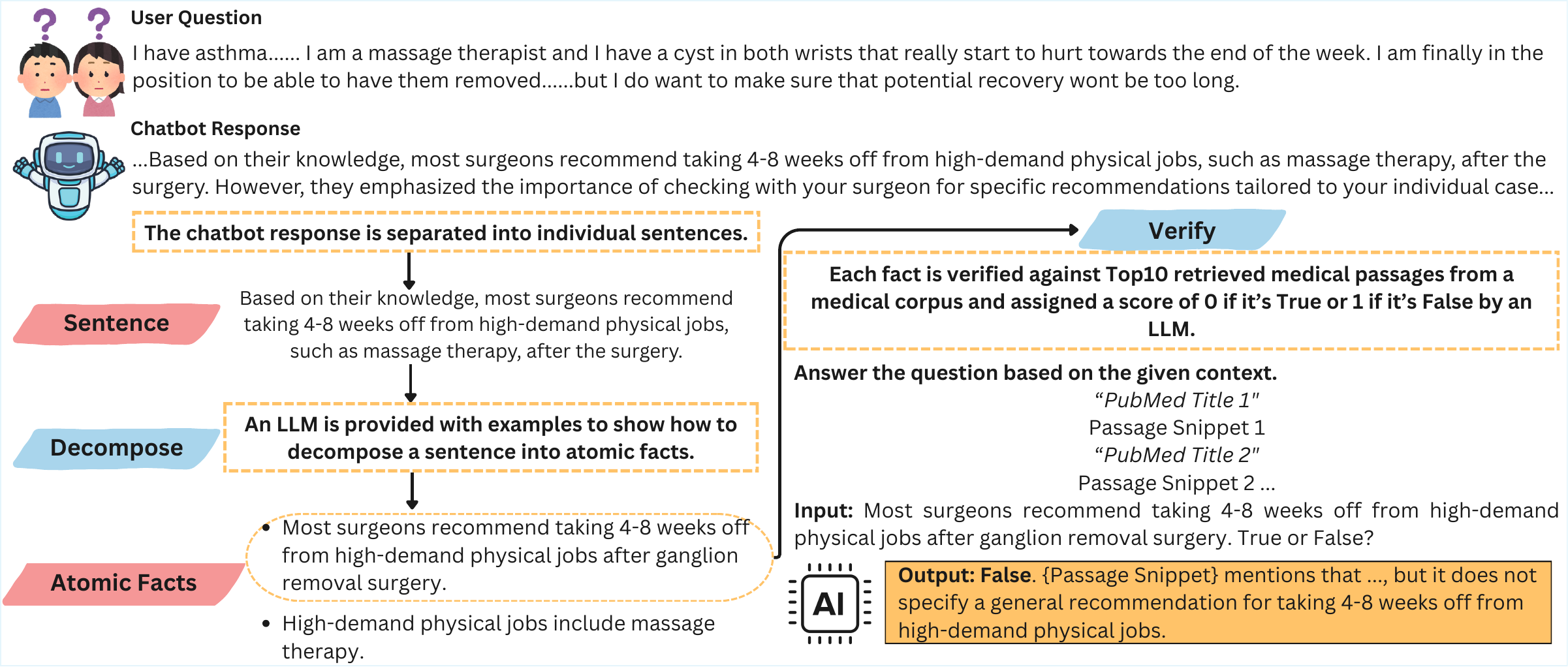}
    \caption{The \textit{decompose-then-verify} pipeline for factuality evaluation on \datasetname, using \methodname\ condition-aware decomposition and medical corpus verification. The \textit{decomposition} step breaks down sentences into one or more ``atomic facts''. The \textit{verification} step checks the factuality of each fact given a context. The context shown in the figure consists of medical passages retrieved from an external medical corpus. The full \datasetname\ data example used here is in \Cref{tab:AskDocsAI_example}. }
    \label{fig:decompose_verify_overview}
\end{figure*}

\section{\taxonomy: Issues with Medical Claim Decomposition}
\label{sec:taxonomy}

Current factuality evaluators,  like FActScore, face several challenges in the medical domain, particularly for patient-facing generations. In contrast to domains for which FActScore and similar systems have been developed, the medical domain has context-dependent, subjective, and highly complex structured sentences.

We describe these challenges in \taxonomy, a framework for analyzing claim decompositions for decompose-then-verify systems. In our analyses, we focus on \textit{invalid}\footnote{\citet{song-etal-2024-veriscore} provide reasons for why a claim can be ``unverifiable'', but the reasons are general and not formalized.} decompositions for medical answers, with examples in \Cref{tab:PUMA_NLI_example}.

In the decompose-then-verify pipeline, LLMs are provided a single sentence for decomposition into atomic claims (\Cref{fig:decompose_verify_overview}). A claim is \textbf{invalid} if it is:

\begin{itemize}[itemsep=1pt]
    \item \textbf{Unverifiable.} While \citet{song-etal-2024-veriscore} define a ``verifiable'' claim as one which ``describes a single event... with all necessary modifiers'', we find it unsuitable for medical answers and use a different definition. An unverifiable claim includes event narratives (``I spoke with your doctor'', ``you are experiencing pain'') or present for patient-empathy (i.e., ``bedside manner'' such as ``Your pain can be very tiring''). These statements are specific to the patient interaction and can't be verified through an external objective knowledge corpus.
    \item \textbf{Hallucinated.} All claims should be grounded in the original sentence with no additional information. A hallucinated claim distorts the meaning of the original sentence or is irrelevant to or adds information not mentioned in the original sentence.
    \item \textbf{Incomplete.} An incomplete claim is overly decomposed and loses the original dependency condition or modifier, which distorts the claim's original meaning. 
    \item \textbf{Incorrectly structured.} A claim not in a declarative format, such as an imperative (e.g. ``Take ibuprofen for your pain''), is hard to verify because it is not in a similar format of sentences in the external authoritative corpus. Also, if a valid claim is nested as a sub-clause (e.g. ``They said \{claim\}''), the wrong prefix will disturb verification judgment for the valid information. These claims need further structure transformation to a declarative format to reduce the verification reasoning burden for LLMs.
    \item \textbf{Context-dependent.} A claim that requires additional information from the context to be stand-alone is context-dependent. This includes claims with vague references to nouns with under-specified or overly-specified pronouns, entities, locations, and temporal information that are difficult or impossible to verify. We reproduced DnDScore, a claim decontextualization method, on \datasetname\ in \Cref{sec:DnDappendix} to find it unadaptable to medical domain text.
    
    \item \textbf{Redundant.} A redundant claim minimally modifies a complete claim without changing the main information. Repeated claims can artificially raise or lower the factuality score of a passage, since the same claim would be evaluated multiple times \citep{wanner-etal-2024-closer}. Often, a redundant claim is also incomplete.
    \item \textbf{Omitted.} The above taxonomy only categorizes generated claims (i.e., precision), while a holistic quality evaluation should also ensure that all ``important'' information from a passage is covered. The definition of ``important'' changes based on the applied domain. We evaluate overly-omitted claims with a ``0-claim'' rate in \Cref{sec:results}.
\end{itemize}

\input{tables/PUMA_NLI_example}

\subsection{Related Work: Current Factuality Evaluation Methods}
\label{sec:related_work}

Factuality evaluation often uses decompose-then-verify multi-agent systems, specialized models, and LLM-as-a-Judge. Specialized models like UniEval \citep{zhong-etal-2022-towards} and Prometheus 2 \citep{kim-etal-2024-prometheus}, and LLMs like GPT-4 \citep{openai_gpt-4_2024} can be used to judge the ``factuality'' of a passage. We focus on the decompose-then-verify systems due to their modularity and ability to identify specific portions of a passage that are non-factual.

The decompose-then-verify system that was popularized by FActScore \citep{min-etal-2023-factscore} has been iterated on by other systems \citep{song-etal-2024-veriscore, jiang_core_2024, huang-chen-2024-factalign, wei_long-form_2024}. While the iterations introduce improvements to the decomposition stage, not all changes are best suited for the medical domain. The taxonomy above identifies issues present in FActScore, some of which have been fixed in subsequent work.
Core introduced a filtering step between the decomposition and verification steps, which removes duplicate and unfaithful claims through measuring entailment. The filtering aims to address \textbf{redundant} and \textbf{hallucinated} claims. Filtering claims implies that there are valid claims as well, which is often not the case with FActScore in the medical domain. Another iteration, VeriScore, introduced a context-inclusive prompt for the decomposition step, which addressed \textbf{context-dependent} and \textbf{incomplete} claims. Also, they fine-tuned a model for the decomposition step, largely solving the \textbf{hallucinated}, \textbf{incorrectly structured}, and \textbf{redundant} issues as well. The main issue with applying VeriScore to medical responses is the high level of \textbf{omitted} claims, as VeriScore decomposes a response into drastically fewer claims than other systems, shown in \Cref{sec:results}.

The factuality of LLMs in the medical domain is often measured as ``correctness'' on multiple-choice medical questions \citep{singhal_towards_2023}. While multiple-choice benchmarks can measure LLM medical knowledge, hallucinations occur only in free responses, which can introduce harmful information to patients. In \Cref{sec:method}, we propose \methodname, a new domain-adapted decompose-then-verify system designed for open-ended medical answers, which addresses all these issues and generalizes well to non-medical open-ended answers. We compare claims from \methodname\ with those from 3 existing decomposition systems, FActScore, VeriScore, and DnDScore, and a claim-filtering system, Core.

\section{\methodname: Domain-adapted Decomposition and Verification Pipeline}
\label{sec:method}

We propose \methodname, a medical text factuality evaluation pipeline consisting of a new decomposition method and domain-adapted verification, designed for open-ended medical answers. We use this decomposition method to transform sentences into atomic claims that can then be verified against different knowledge corpora. 
We consider three verification methods that differ by the source for claim-level verification: model internal parametric knowledge, a customized reference corpus, and a medical retrieval corpus. 
Figure \ref{fig:decompose_verify_overview} shows the decompose-then-verify framework for a single sentence in a chatbot response. 

To support our system development and evaluation, we create a new free-response medical QA dataset, \datasetname. We also evaluate \methodname\ on an existing medical QA dataset with a greatly different writing style, PUMA, and on a culturally specific Long-Form Question Answering (LFQA) dataset, CaLMQA, to test system robustness in both medical and non-medical domains.

\subsection{Medical and Non-Medical Free-Form Answer Data}
\label{sec:data}
We evaluate \methodname\ using two medical and one non-medical free-form question answering datasets. Detailed dataset statistics are in \Cref{tab:dataset-stats}.

\textbf{\datasetname}
Following the work comparing physician and chatbot responses to patient questions \citep{ayers-comparing}, 
we selected 300 medical questions with long responses from the r/AskDocs\footnote{\url{https://www.reddit.com/r/AskDocs}} subreddit, where verified doctors respond to user questions. Rather than generating a novel response to the question, we prompted an LLM with both the question and doctor response pair and asked the model to augment the doctor response by explaining unclear terms and including empathic statements without changing the doctor's recommendations, e.g., introducing new diagnoses or treatments. We generated chatbot responses using \texttt{Meta-Llama 3.1-405B Instruct-Turbo} \citep{dubey2024llama3herdmodels} to ensure high-quality generations. The detailed data creation process is in \Cref{sec:AskDocs Creation}. 

\textbf{PUMA} is a health question-answer dataset filtered from samples on L6 - Yahoo! Answers Comprehensive Questions and Answers. PUMA contains 3,195 user questions with 10,067 answers (roughly 3 answers per question), and manually annotated answers' text spans with 5 labels: Cause, Suggestion, Experience, Question, and Information. We treat each answer-question pair as an individual example. Unlike \datasetname, the answers are neither written by doctors nor rewritten by an LLM, and are generally shorter and of lower quality. Since we developed \methodname\ on \datasetname, we consider PUMA an ``out of domain'' test set for robustness evaluation in \Cref{sec:PUMA_generalize}.

\textbf{CaLMQA} is a cultural free-form QA dataset\footnote{We used CaLMQA's initial 2024 version on Huggingface, but it was updated on June 12, 2025, with some questions removed. We kept the English subset of it on GitHub for our result reproducibility.}, whose 96 English QA pairs are collected from r/explainlikeimfive\footnote{\url{https://www.reddit.com/r/explainlikeimfive}} (ELIF) subreddit. We showed that only minimal changes to \methodname\ are required to adapt it to a non-medical domain in \Cref{sec:cultural_generalize}.

\subsection{Decomposition}
The FActScore few-shot decomposition method was based on biographical data, which differs from the structure of medical chatbot responses.

Our \methodname\ decomposition prompt includes detailed instructions to describe and solve the corresponding deficiencies we outlined in \taxonomy\ in \Cref{sec:taxonomy}, and append 10 high-quality examples after the instructions. At the end of the prompt, we insert the sentence to be decomposed with its context to nudge the model to output extracted facts in the sentence. We put details of this new taxonomy-based decomposition prompt creation process in \Cref{sec:MedScore_decomp_creation}, and the full prompt is in \Cref{tab:MedScorePrompt}. We use GPT-4o-mini \citep{openai2024gpt4ocard} for claim decompositions. The ablation study to assess the impact of each component in the \methodname\ decomposition prompt is in \Cref{sec:ablation_study}.

\subsection{Verification}
\label{sec:verify}

After \textit{decomposing} the sentences into atomic claims, we \textit{verify} the claims separately against three sources: LLMs' internal knowledge, the original doctor response (only applicable to \datasetname), and an external medical corpus, MedCorp. 

\textbf{Internal Knowledge.}\footnote{Models' Internal Knowledge is also called Parametric Knowledge in other work} We prompt a close-sourced general-purpose model, \texttt{GPT-4o}, an open-sourced general-purpose model, \texttt{Mistral-Small-24B-Instruct-2501} (Mistral Small 3) \citep{mistral_ai_mistral_2025}, and an open-sourced medical-specialized model, \texttt{Llama3-OpenBioLLM-70B} (OpenBioLLM) \citep{OpenBioLLMs}, to use their own knowledge to determine whether a claim is True or False. The internal knowledge test prompt is in Appendix \Cref{tab:internal-prompt}.

\textbf{Original Doctor Response.} We use FActScore's method to prompt Mistral Small 3 for claim verification against the original doctor response. We provide the verifier model with a claim and its corresponding original doctor response as the context. The verifier model is asked if the claim is ``True or False'' given the context, with the verification prompt in Appendix \Cref{tab:verification_prompt}.

\textbf{External Medical Corpus.} We used the MedRAG retrieval system \citep{xiong-etal-2024-benchmarking}, along with the MedCorp dataset, to retrieve relevant medical literature passages to evaluate the claims. MedCorp consists of four corpora: PubMed, StatPearls, Textbooks, and Wikipedia\footnote{After qualitative evaluation, we removed Wikipedia from the corpus due to the occasional retrieval of medical-related TV and movie plot points.}, and the ablation study for each corpora's impact is in \Cref{sec:ablation_medcorp}. For each claim, we retrieve the top-10 most relevant passages 
ranked by a medical retriever, MedCPT \citep{jin_medcpt_2023}, with the top-$k$ hyperparameter selection experiment in \Cref{sec:topk_selection}.
As in \Cref{fig:decompose_verify_overview}, these passages are provided as the \textit{context} for the verification prompt in \Cref{tab:verification_prompt}. 

A True claim is given a score of $1$ and a False claim has a score of $0$. The factuality score for a response is its average claim score between $[0.0, 1.0]$. 
We selected Mistral Small 3 as the verifier model for the Doctor Response and External Medical Corpus experiments after conducting a model comparison experiment in \Cref{sec:verifier-selection}.  

\subsection{Comparative Systems}
We compare \methodname\ decomposition strategy to existing decomposition methods, including \textbf{FActScore} (few-shot), \textbf{VeriScore} (supervised finetuning), and \textbf{\methodname+Core} (post-filtering), with additional comparison to \textbf{DnDScore} (explanation-based prompting) in \Cref{sec:DnDappendix}. Baseline implementation details are in \Cref{sec:baseline_details}.

\section{Results}
\label{sec:results}
We provide quantitative and qualitative analyses by comparing the claims decomposed from \methodname, FActScore, VeriScore, and \methodname+Core according to \taxonomy\ in \Cref{sec:results_quality}.
We also explore the impact of verification method on the final factuality score in \Cref{sec:results_verify} and the robustness of the decomposition methods on two out-of-domain datasets, PUMA and CaLMQA in \Cref{sec:generalize_results}.
The experiment computation and hyperparameter details for reproducibility are in \Cref{sec:compute}. 

\begin{table*}[h]
    \centering
    \resizebox{\textwidth}{!}{
    \begin{tabular}{llcccc}
        \toprule
        & & FActScore & \methodname & VeriScore & \methodname+Core \\
        \cmidrule{3-6}
        \multirow{3}{*}{\datasetname}
        & 0-claim rate & 0\% & \textbf{0\%} & 14.67\% & 0.3\%\\
        & \#claims/response & 28.60 (9.3) & 11.94 (4.4) & 3.87 (3.1) & 7.92 (3.3)\\
        & \#claims/sentence & 4.24 (1.2) & 1.77 (1.2) & 0.57 (1.0) & 1.17 (1.1)\\
        \midrule
        \multirow{3}{*}{PUMA}
        & 0-claim rate & 0\% & \textbf{9.3\%} & 53.69\% & 27.6\% \\
        & \#claims/answer & 13.30 (17.6) & 6.68 (11.2) & 2.49 (6.2) & 3.43 (5.8) \\
        & \#claims/sentence & 3.54 (1.7) & 1.78 (1.7) & 0.68 (1.4) & 0.92 (1.2) \\
        \midrule
        \multirow{3}{*}{CaLMQA}
        & 0-claim rate & 0\% & \textbf{0\%} & 7.29\% & 2.08\% \\
        & \#claims/answer & 22.60 (18.2)  & 13.50 (11.2) & 8.25 (7.1) &  7.62 (6.9)\\
        & \#claims/sentence & 3.81 (1.4) & 2.28 (1.3) & 1.30 (1.2) & 1.26 (1.2)\\
        
        \bottomrule
    \end{tabular}
    }
    \caption{Claim statistics by decomposition method on \datasetname, PUMA, and CaLMQA. The numbers reported are averaged over the dataset, and the standard deviation is shown in parentheses.}
    \label{tab:combined_claim_stats}
\end{table*}

\begin{table*}[h]
    \centering
    \begin{tabular}{l|c|c|c|c}
    \toprule
        The number of & FActScore & \methodname & VeriScore & \methodname+Core\\
    \midrule
        Total claims & 212 & 86 & 24& 60\\
        Valid claims & 36 (17\%) & \textbf{64 (74.4\%)} & 22 (91.7\%) & 42 (70\%) \\
        Unverifiable claims &  79 (37.3\%) & 8 (9.3\%) & 1(4.2\%)& 8 (13.3\%)\\
        Hallucinated claims & 15 (7\%) & 1 (1\%) & 1 (4.2\%) & 0\\
        Incomplete claims & 30 (14.2\%) & 9 (10.5\%)& 0 & 7 (11.7\%)\\
        Wrong-structure claims & 26 (12.3\%) & 4 (4.7\%) & 0 & 2 (3.3\%) \\
        Context-dependent claims & 50 (23.6\%) & 3 (3.5\%) & 0 & 3 (5\%)\\
        Redundant claims & 27 (12.7\%) & 0 & 0 & 0\\
    \bottomrule
    \end{tabular}
    \caption{Human evaluation of invalid claims across decomposition methods for 10 \datasetname\ samples (IAA=0.73).}
    \label{tab:wrong-claim-rate}
\end{table*}

\subsection{Claim Quality Evaluation}
\label{sec:results_quality}

The claim statistics on the response- and sentence-level for all decomposition strategies are in \Cref{tab:combined_claim_stats}.\footnote{See \Cref{sec:claim_decomp_details} for additional visualization figures.} For the \datasetname\, we assume every response has at least one valid claim since the chatbot responses are high-quality, rewritten doctor responses. When a decomposition method returns no claim for a response, this indicates that there is no verifiable claim in the response, which contradicts the fact that every response usually has at least one piece of verifiable medical information. Therefore, we consider the ``0-claim rate'' to be an indicator of \textbf{omitted} claims, and it is a rough estimation of the claim recall and coverage quality.

In addition to the quantitative analysis, we randomly sampled 10 responses to label their 382 claims with the previously defined 6 error categories in \taxonomy\ in \Cref{sec:taxonomy} for manual qualitative analysis. The annotation guidelines are in \Cref{sec:annotation}. The annotation was performed by two annotators with an overall inter-annotator agreement of 0.73 in Cohen's kappa. The rounded average results are in \Cref{tab:wrong-claim-rate}.
In general, \textbf{\methodname\ has the most number of valid claims with low error rates, highlighted in bold, among the four decomposition methods.}

\begin{table*}[b]
    \centering
    \resizebox{\textwidth}{!}{
    \begin{tabular}{llccccc}
    \toprule
    Dataset & Decomposition & GPT-4o & OpenBioLLM & MistralSM & Dr. Response* & Retrieved Docs \\
    \cmidrule{3-7}
    \multirow{3}{*}{\datasetname}
    & FActScore  & 31.96\% & 84.32\% &  46.10\% & 80.94\% & 50.04\% \\ 
    & \methodname & 76.54\% & \textbf{95.64\%} & \textbf{78.09\%}& \textbf{94.68\%} & \textbf{70.07\%} \\
    & VeriScore & 81.94\% & 94.32\% & 77.23\% & 94.41\% & 68.16\% \\ 
    \midrule
    \multirow{3}{*}{PUMA}
    & FActScore & 35.35\% & 80.85\% & 40.55\% & - & 47.34\% \\
    & \methodname & \textbf{78.54\%} & \textbf{92.82\%} & \textbf{71.31\%} & - & \textbf{61.56\%} \\
    & VeriScoreQA & 77.98\% & 90.95\% & 65.94\% & - & 59.46\% \\
    \midrule
    \multirow{3}{*}{CaLMQA}
    & FActScore & 39.82\% & - & 43.69\% & - & 60.42\% \\
    & \methodname & 70.83\% & - & 62.60\% & - & \textbf{68.96\%} \\
    & VeriScoreQA & \textbf{74.83\%} & - & \textbf{63.44\%} & - & 62.22\% \\

    \bottomrule
    \end{tabular}
    }
    \caption{Factuality scores of \datasetname\ chatbot responses and the PUMA, CaLMQA answers verified against model internal knowledge (GPT-4o, MistralSM, OpenBioLLM), and retrieved documents. We use MedCorp as the corpus for \datasetname\ and PUMA, and Wikipedia for CaLMQA. The Dr. Response from \datasetname\ is only used to evaluate the quality of machine augmented answers in the \datasetname\, and is not used for factuality evaluation.}
    \label{tab:verification_results}
\end{table*}

\textbf{FActScore over-generates claims.} From \Cref{tab:combined_claim_stats}, we find that FActScore generates the most number of claims among the three decomposition methods, but it overly decomposes sentences into invalid claims, whose overall valid claim rate is only 17\% in \Cref{tab:wrong-claim-rate}.

\textbf{VeriScore and Core over-filter claims.}
\Cref{tab:combined_claim_stats} shows that VeriScore generates 0 claims for 44 responses and its 0-claim rate is 14.67\% for 300 long responses in \datasetname. \Cref{tab:wrong-claim-rate} shows that it generates the fewest number of invalid claims, which indicates that VeriScore sacrifices the number of claims for claim quality, omitting a lot of informative facts that need to be verified. Additionally, the post-filtering system, Core, can not distinguish valid and invalid claims and equally filters them out by 33\% on \datasetname, 49\% on PUMA and 44\% on CaLMQA.

\textbf{Decomposition methods decompose to different ``atomic or molecular'' fact levels.}
The average number of tokens of claims extracted by FActScore, \methodname, and VeriScore on \datasetname\ is 10.31, \textbf{13.03}, and 12.77, respectively. These claim lengths show that FActScore decomposes sentences into atomic facts while \methodname\ and VeriScore generate molecular facts by including more information from the given context.

\subsection{Factuality Scores against Different Corpora}
\label{sec:results_verify}
The truthfulness of a claim varies by the corpus we verify it against. We evaluate three different verification corpora in \Cref{sec:verify}: LLM internal parametric knowledge (\texttt{GPT-4o}, Mistral Small 3, and  OpenBioLLM), doctor responses, and retrieved medical passages from MedCorp. 
We first excluded all 0-claim responses that were incorrectly decomposed to have no verifiable information by each decomposition method, and then calculated the average factuality score of the remaining responses. 
The results are in \Cref{tab:verification_results}. We also manually examine the verifier models' output to find the reasons for the factuality score variance.

\textbf{FActScore underestimates and VeriScore overestimates the factuality score.} Given the same chatbot responses, the evaluation of the factuality score varies substantially by decomposition method. FActScore consistently leads to the lowest factuality score, regardless of the verification corpus, which again proves its low-quality claim generation. VeriScore has similar or slightly higher factuality scores than \methodname\ because, although they have similar quality claims, VeriScore has much fewer claims in total, resulting in smaller denominators and the final higher quotients, overestimating the factuality scores. FActScore and VeriScore can not be used off-the-shelf, while our customizable pipeline can be adapted to out-of-domain datasets with minimal effort, as shown in \Cref{sec:generalize_results}, for more reliable score estimation.

\textbf{General-purpose LLMs are more robust in using their own knowledge to identify invalid claims than medical-specific open-sourced LLMs.} Compared to OpenBioLLM, a model fine-tuned for medical domain tasks, GPT-4o and Mistral Small 3 are more robust to identifying invalid claims as False, as shown by the 45-52\% lower factuality scores for FActScore on two medical datasets. They are specifically better at identifying \textbf{unverifiable} claims. OpenBioLLM can not distinguish valid and invalid claims well, thus overly assigning True to most invalid claims and resulting in inflated scores.

\textbf{\datasetname\ machine augmented answers are faithful to the original doctor responses.} When we verify the chatbot responses' claims against doctor responses, the factuality scores can achieve 94\%, indicating that the machine augmented answers are high quality because they are grounded in the doctor response with minimal hallucinations. From manual analysis, we notice that many \textbf{unverifiable} claims, e.g., event narratives, are verified as True because they are present in or entailed by the doctor's response. Therefore, its factuality scores are inflated by True but \textbf{invalid} claims, so it can only be used to evaluate how much information from the chatbot responses is grounded in the original doctor response. 

\textbf{Claims are harder to verify against external corpora.} Verifying against the internal knowledge of LLMs assigns higher factuality scores to claims than when verifying against the external MedCorp. The factuality scores by \methodname\ and VeriScore drop by 26\% at most, to 68-70\% on \datasetname. In the external corpus verification, we find that Mistral Small 3, the backbone verifier LLM, has strong capabilities to understand medical passages, associate the claim with the passages, and reason about the truthfulness of the claim. However, verification also directly relies on the quality of the corpus retriever and on whether any supporting information is present in the corpus. The lack of relevant passages adds an additional challenge to the retriever to find evidence to support a claim, which is the main reason for false negatives (i.e., the claim is True but the verifier assigns False). Another issue is false positives, which happens when a too-general claim (typically \textbf{incomplete} or \textbf{context-dependent}) is highly relevant to many corpus documents, and is assigned True by the verifier, but should be False for that specific case.

\subsection{Generalizability on Out-of-Domain Data: PUMA and CaLMQA}
\label{sec:generalize_results}
We developed and tested our prompts and system only on \datasetname. Therefore,  we treat PUMA and CaLMQA as two out-of-domain datasets to validate our system. We find the same patterns as in \Cref{sec:results_quality} for decomposition quality and in \Cref{sec:results_verify} for verification, showing that our system is generalizable to both medical and non-medical domains.

\subsubsection{Medical domain generalizability}
\label{sec:PUMA_generalize}
Without changing the decomposition prompt of \methodname, we decompose an answer in PUMA using \texttt{GPT-4o-mini} and use the answer itself as the answer context, and combine its corresponding user question and the question context as the full question context for the LLM to find necessary information. To produce VeriScore's highest quality claims for comparison, we use its released QA prompting method to extract claims and name it as VeriScoreQA hereafter.

PUMA also contains manual annotations, which we can use to evaluate the quality of the decomposed claims. As described in \Cref{sec:data}, PUMA answers have text spans labeled with 5 categories: Cause, Suggestion, Experience, Question, and Information. According to \taxonomy, only claims that are decomposed from sentences in the Cause, Suggestion, and Information spans can be considered \textbf{verifiable} claims. Experience and Question spans describe users' first-hand experiences or questions, which are usually subjectively unverifiable and don't need to be verified.

Therefore, we prompt \texttt{gpt-4o-mini} as a Natural Language Inference (NLI) model to identify which claims can be inferred by a labeled text span. The NLI prompt is in Appendix \Cref{tab:NLI_prompt}.

An answer in PUMA is verifiable if it contains at least one text span from a verifiable category (e.g., Cause). The verifiable rate of an answer is the ratio of the claims that can be inferred by its verifiable spans to all its claims, which is in [0.0, 1.0]. 

While the verifiable rate is akin to precision, the recall of verifiable claims can be approximated by the complement of the 0-claim rate on verifiable answers. Specifically, if no claims were decomposed from a verifiable answer, the decomposition method should be penalized for \textbf{omitting} claims. To provide a holistic view of the claim decomposition quality, we penalize a verifiable answer with 0 claim by a score of -1 and calculate the adjusted verifiable rates in \Cref{tab:PUMA_NLI_result}. 

\input{tables/PUMA_NLI_result}

\textbf{\methodname\ includes the most verifiable information with minimal inclusion of invalid claims.} VeriScoreQA has the highest verifiable claim rate in \Cref{tab:PUMA_NLI_result} but has the lowest overall quality by the adjusted verifiable rate after 0-claim penalization, due to omitting too many verifiable answers. \methodname's overall verifiable rate slightly changes by 1.4\% after penalization, compared with VeriScoreQA's 35\% score drop, which shows its great performance in both retaining verifiable answers and omitting unverifiable answers. FActScore isn't influenced because it always generates claims and has no 0-claim answers. However, the verifiable rate can only indicate if a claim contains the information from verifiable spans, but can not indicate its quality (i.e., validity defined by \taxonomy). Although FActScore's verifiable rate is high, its claims are still invalid by containing lots of erros in the \taxonomy, such as incomplete, context-dependent, and redundant, shown in \Cref{tab:PUMA_NLI_example}. \textbf{In general, \methodname\ has the highest verifiable span claim coverage by the least omission penalization and the highest adjusted verifiable rate in \Cref{tab:PUMA_NLI_result}.}

Some \methodname\ claims are verifiable and valid, but there are no corresponding annotated verifiable spans in the PUMA dataset, which results in an underestimate of its verifiable rate, as shown in \Cref{tab:FalseNegative_NLI_example}. 
\input{tables/False_Negative_NLI_examples}

\textbf{\methodname\ is more robust to colloquial language.}
Similar to the chatbot response decomposition results, FActScore has the most claims while VeriScoreQA has the fewest claims in \Cref{tab:combined_claim_stats}. The 0-claim rate of VeriScoreQA becomes extremely high, 53.69\%, because PUMA answers are shorter, full of colloquial sentences and slang (e.g., ``LOL''), and have less information density. \methodname\ has a more reasonable 0-claim rate of 9.3\% since some answers are too short (e.g., single-word answer: ``powder.'') and contain no verifiable information. 
The average number of tokens of claims extracted by FActScore, \methodname, and VeriScoreQA is 9.38, \textbf{11.39}, and 9.33, respectively, showing that \methodname\ generally retains more information in the claims from the input.

When the verification corpus is the same, invalid claims are the main reason for low factuality scores. \Cref{tab:verification_results} shows that \methodname\ achieves the highest factuality scores among all verification corpora\footnote{Due to cost, we sampled 1000 answers from PUMA for the GPT-4o internal knowledge test.}, showing its robustness to the style of writing in free-form answers by generating more valid claims than other methods.

\subsubsection{Non-medical domain generalizability}
\label{sec:cultural_generalize}

To show \methodname\ non-medical domain performance, we selected a culturally specific Long-Form Question Answering (LFQA) dataset, CaLMQA, to test the non-medical domain robustness. The English LFQA data were collected from Reddit ELI5 and contain 96 QA pairs. We only made minimal changes to the \methodname\ to adapt to a different domain:

1. \textbf{Modify the ICL prompt and remove medical-specific wording.} Everything in the \methodname\ decomposition prompt is the same, except we changed two words in the first sentence from ‘You are a medical expert in evaluating how factual a medical sentence is.’ to ‘You are an expert in evaluating how factual a sentence is.’ to adapt to the non-medical setting.

2. \textbf{Use an appropriate verification corpus.} We changed the verification corpus from the medical corpus to Wikipedia and changed the retriever from the medically fine-tuned retriever to a general domain retriever, Contriever \citep{izacard2022unsupervised}.

\input{tables/CalmQA_examples}

From the result table in \Cref{tab:verification_results}, we find the same pattern: FActScore overly decomposes sentences into invalid claims, and VeriScore overly omits claims that should be generated and verified. \textbf{MedScore generates twice as many claims as VeriScore and is not specifically tuned for the best performance on non-medical data, while VeriScore has fine-tuned its model on data from the same Reddit ELI5 to optimize performance.} Tuning-free \methodname\ includes the most verifiable information with minimal inclusion of invalid claims, as shown in \Cref{tab:CaLMQA_example}.

In terms of the factuality score, MedScore shows the least score variance (1.87\%) by the two verification corpora (GPT-4o and WikiRAG), than FActScore (20.6\%) and VeriScore (12.61\%), indicating its stability in approximating the true fine-grained factuality score, which is unknown because open-ended answers don’t have gold references.

\section{Conclusion}

We proposed \methodname, which decomposes medical answers into condition-aware valid claims while preserving most of the informative content that needs to be verified and verifies claims against domain-adapted knowledge corpus. \methodname\ decomposition method generalizes well to out-of-domain free-form data and can be modified easily to adapt to other fields' text and customized verifiability definitions, using the released modular domain-adaptable pipeline. 
Given that existing popular systems, FActScore, VeriScore, and DnDScore, should not be used off-the-shelf, our paper's workflow serves as guidance for researchers to modify our pipeline with minor effort for domain adaptation. In summary, customized decomposition, appropriate verification corpora, and strong capacity LLMs can provide a more reliable estimate of the true factuality score.

\section*{Limitations}
\paragraph{MedScore measures accuracy, but not completeness.}
Regarding medical response accuracy, the decompose-then-verify framework cannot measure missing information, which means the completeness evaluation of the answers. It only verifies the information that is present (i.e., precision). However, to ensure the safety of medical chatbot applications, we should evaluate whether a medical answer includes all the necessary information for a patient. In terms of real-world free-form QA, there are no gold standard answers that can be seen as fully complete. Therefore, we don't have gold references to compute information recall in diverse machine-generated answers. This remains a challenge for future work. 

\paragraph{Potential bias in LLM verifiers.}
The fact verification methods presented in \Cref{sec:verify} depend on the LLM's capabilities to follow instructions and comprehend retrieved passages. While LLMs have been shown to excel at these tasks \citep{wang-etal-2022-super,zhong2023agievalhumancentricbenchmarkevaluating}, there is also work demonstrating bias in using LLMs as evaluators \citep{wang-etal-2024-large-language-models-fair,NEURIPS2023_91f18a12}. While this is a concern, model intrinsic bias is outside the scope of this paper.

\paragraph{Limited number of doctor-written free-form answer datasets to patient questions.}
Existing medical datasets or benchmarks mainly focus on multiple-choice QA, instead of patient-facing colloquial language written by doctors. Our primary goal is to assess the medical chatbot response quality, but there are not many \textbf{publicly available} datasets with chatbot responses and doctor gold answers or annotations. AskDocs subreddit is the best publicly available forum where we can collect answers from verified doctors, instead of laypeople. There are multiple widely cited research papers using AskDocs’s responses as gold references\citep{nguyen-etal-2023-medredqa, abrar2024empiricalevaluationlargelanguage, ayers-comparing}, which proves its good quality among public forums. However, we still didn’t choose to use doctor responses as the gold reference for verification. Only medical literature corpus, like PubMed, and LLM internal knowledge are used in verification.

\section*{Ethical Considerations}
Few areas in the greater natural language processing community are as high-risk as patient-facing medical Generative AI systems regarding the sensitivity of data and the need for accuracy. In this work, we present \datasetname, a dataset to support research in generating accurate and safe medical-question free responses. We collected the data from Reddit, a public forum, and do not add any user-specific information beyond what is self-disclosed. 


\input{main.bbl}\bibliographystyle{colm2025_conference}

\appendix
\section{Appendix}
\label{sec:appendix}

\subsection{DnDScore Claims on \datasetname}
\label{sec:DnDappendix}
We reproduced DnDScore \citep{wanner_dndscore_2024} on \datasetname\ and found most of the generated claims are invalid and not sufficiently decontextualized, as shown in \Cref{tab:DnD_example}. DnDScore first generates subclaims for each sentence, using the full response paragraph as the context, and then decontextualizes each subclaim into a decontextualized claim by adding additional information. Its first step generates similar quality claims as FActScore, and the second step focuses on only one error in the \taxonomy, making it fail to create valid claims on medical domain text.
\input{tables/DnD_example}

\subsection{Dataset Statistics}

The statistics of \datasetname\, PUMA, and CaLMQA are in \Cref{tab:dataset-stats}. \datasetname\ has doctor responses and chatbot responses, while PUMA and CaLMQA only have user responses, so there are two N/A in their columns' doctor response rows. CaLMQA doesn't separate questions from context, so it is hard to calculate how many questions have additional context information, and we mark it as N/A. In the \datasetname\ dataset, chatbot responses usually have more and longer sentences than doctor responses. In addition, doctor responses are not themselves complete and might have information omissions due to subjective writing preferences. 

\begin{table*}[h]
    \centering
    \begin{tabular}{lccc}
    \toprule
       (Average) number of & \datasetname & PUMA & CaLMQA\\
    \midrule
        user questions & 300 & 3195 & 96\\
        user questions with context* & 300 & 1173 & N/A\\
        doctor, chatbot/user responses & 300, 300 & 10067 & 96\\
       sentences per user question  & 12.3  & 1.2 & 2.39 \\
       sentences per doctor response  & 5.1 & N/A & N/A\\
       sentences per chatbot/user response & 6.7 & 3.8 & 5.93\\
       tokens per sentence in user question & 18.3 & 10.9 & 39.4\\
       tokens per sentence in doctor response & 18.7 & N/A & N/A \\
       tokens per sentence in chatbot/user response & 24.5 & 15.4 & 123.1\\
    \bottomrule
    \end{tabular}
    \caption{Statistics for \datasetname, PUMA, and CaLMQA datasets. *Some users write additional information as the context of their question, such as their medical history and personal experiences.}
    \label{tab:dataset-stats}
\end{table*}

\subsection{\datasetname\  Data Collection}
\label{sec:AskDocs Creation}
The \datasetname dataset was curated from the r/AskDocs subreddit, which features medical questions from users and responses from verified healthcare professionals. We collected posts spanning from December 31, 2015, to December 27, 2022. To ensure quality and relevance, we applied several reproducible filtering steps by coded rules without human intervention or bias.

First, we identified responses from verified medical professionals by filtering for users with physician-specific flairs (e.g., "Physician," "Physician - Cardiology," "Physician - Dermatologist"). We excluded posts with fewer than 50 words to ensure substantive content and removed posts containing URLs to minimize reference dependencies. For posts with multiple physician responses, we selected either the earliest response or the most prominently marked one (e.g., from a verified specialist or moderator).

Each entry in our dataset contains a unique post ID, the patient's question, the physician's response, metadata on response selection criteria, and a flag indicating the presence of links in the original post. We performed standard preprocessing steps including ASCII encoding, escape sequence normalization, and whitespace standardization to ensure consistency across all entries.

From this initial collection of 343 question--physician pairs, we randomly selected 300 samples for inclusion in the final \datasetname dataset. The selection process prioritized clarity, comprehensiveness, and diversity of medical topics. Importantly, we deliberately avoided using post score (upvotes) as a filtering criterion, as this can introduce bias toward popular but not necessarily more informative content.

For our experiments, we used these 300 physician responses as a gold standard and generated AI responses using Meta-Llama 3.1-405B Instruct-Turbo with prompt in Table \ref{tab:AskDocs_prompt}. The generation settings included a maximum token length of 512, temperature and top-p sampling both set to 0.7, top-k filtering set to 50, and a repetition penalty of 1.0. We generated "augmented responses" by providing the model with both the original patient question and the physician's response, instructing it to make the medical information more accessible to patients without introducing new diagnoses or treatments beyond what the physician had stated. This approach aimed to enhance readability and patient understanding while maintaining the clinical accuracy of the original physician response. \Cref{tab:AskDocsAI_example} has one example of a user question, doctor response, and chatbot augmented response.

\input{tables/AskDocsAI_example}

\subsection{Details on \methodname\ Decomposition Prompt Creation}
\label{sec:MedScore_decomp_creation}

At first, we created 8 diverse decomposition examples with GPT-4o \citep{openai2024gpt4ocard} using the FActScore in-context learning (ICL) prompt.
The first 8 ICL examples are drawn from 8 randomly sampled responses, in the order of the first sentence in the first response, the second sentence in the second response, …, the last sentence in the last response. Then we manually checked, replaced wrong claims, and added new examples to make sure all ICL examples include at least one example for each invalid category we defined in \Cref{sec:taxonomy}. For example, there should be at least one example that decomposes a claim with the condition in the original sentence and at least one example that transforms imperative sentences, "take ibuprofen for pain", into declarative format "ibuprofen is helpful for pain". We finalized our ICL example in \Cref{tab:MedScorePrompt} by the ablation study in \Cref{sec:ablation_study}.

\subsection{Ablation Study for \taxonomy-based Decomposition Prompt}
\label{sec:ablation_study}
To assess the impact of each component in the \methodname\ decomposition prompt on the overall decomposition quality, we conducted ablation studies across six settings: (1) 8 ICL examples only, which were created from the process in \Cref{sec:MedScore_decomp_creation}, in \Cref{tab:MedScore_Ablation_Prompt}, (2) Taxonomy-based Instruction only, (3–6) Taxonomy-based Instruction with the first 4 ICL, 8 ICL, 8 balanced ICL (4 positive, 4 negative), and 10 balanced ICL (5 positive, 5 negative) examples in \Cref{tab:MedScorePrompt}. We manually examined the generated claims and found that the taxonomy-based instruction with 10 balanced ICL examples has the best quality. We verified the claims against GPT-4o internal knowledge and external medical corpus. Results in \Cref{tab:ablation_result} show that LLMs can not learn valid claim patterns only from 8 ICL examples, producing similar quality claims to FActScore. Taxonomy-based Instruction most effectively guides LLM to avoid generating invalid claims, but often leads to plausible yet inference-based outputs not strictly grounded in the sentence. Adding 4 ICL examples helps regulate format and alignment with examples. However, simply increasing the number of ICL examples from 4 to 8 does not improve the claim quality. We observed that the 3:1 positive-negative ratio in the 8 ICL examples limited performance. Switching to a 1:1 ratio with sufficient ICL examples in \Cref{tab:MedScorePrompt} yielded the best results.

\begin{table*}
    \centering
    \resizebox{\textwidth}{!}{\begin{tabular}{l|c|c|c|c|c|c}
    \toprule
         & ICL\_Only & Inst\_Only & Inst+4ICL & Inst+8ICL & Inst+8BalICL & Inst+10BalICL\\
    \midrule
       0-claim rate & 0\% & 0\% & 0\% & 0\% &0\% &0\%  \\
       \#claims/response & 18.15 (6.0)	 & 13.66 (5.0) 	&13.11 (4.7)	&13.22 (4.7)	&12.28 (4.44)&	11.94 (4.43)\\
       \#claims/sentence & 	2.69 (1.0)&	2.02 (1.5)	&1.94 (1.1)	&1.96 (1.2)	&1.82 (1.1)&	1.77 (1.2)\\
        \#tokens/claim&	12.20	&12.55	&12.87	&12.95	&13.06	&13.03\\
        GPT-4o score&44.06\%	&82.38\%	&73.57\%&	71.92\%&	75.39\%&	\textbf{76.54\%}\\
        MedRAG score	&47.73\%&	73.65\%&	65.74\%	&64.25\%	&66.47\%	&\textbf{66.88\%}\\
    \bottomrule
    \end{tabular}}
    \caption{Ablation study result of \methodname\ decomposition prompt. The numbers reported are averaged over the \datasetname\ dataset, and the standard deviation is shown in parentheses. MedRAG score is based on Top-5 retrieved passages.}
    \label{tab:ablation_result}
\end{table*}

\subsection{LLM Prompts for \methodname\ Pipeline}
The decomposition prompt of \methodname\ is in \Cref{tab:MedScorePrompt}. The verification prompt for internal LLM knowledge is in \Cref{tab:internal-prompt} and the one used for the original doctor responses and retrieval-based verification is in \Cref{tab:verification_prompt}.

\subsection{Top-k Parameter Selection for Retriever}
\label{sec:topk_selection}
We rank the retrieved passage snippets by their relevancy score to the claim we want to verify. The score is calculated by the cosine similarity of the embeddings of a snippet and a claim. Since too many passages will slow down and perturb the verification step greatly, and too few passages will lead to a lack of evidence issue, we want to select the optimal number (top-k) of snippets by balancing the cost and effect. The running time analysis result for the number of retrieved snippets with \methodname\ decomposition on \datasetname\ is in \Cref{tab:topk_selection}. We find that increasing the number of snippets can improve the factuality score by 0.5\%-3\%, because over 70\% of the False claims in the top-5 setting are due to a lack of direct evidence in the retrieved snippets, as discussed in \Cref{sec:results_verify}. However, as the relevancy decreases, the benefit of including more snippets decreases, and the running time increases much more. Considering the tradeoff for efficiency, selecting the top 10 snippets is the optimal choice in this paper. 
\input{tables/TopK_selection}

\subsection{Ablation Study for MedCorp Individual Corpus}
\label{sec:ablation_medcorp}
\input{tables/MedCorp_ablation}
We used MedScore decomposition on AskDocsAI and retrieved the top 5 passages by relevancy score for the MedRAG ablation study, and the results are in \Cref{tab:medcorp_ablation_result}. We find PubMed, the largest corpus, takes the longest time to retrieve and results in the lowest score. StatPearls and Textbooks are much smaller, thus quicker, and have 2\%-3\% higher scores than PubMed. However, within the retrieved snippets, PubMed has the largest number of relevant snippets to a claim, more than Statspearls and Textbooks. In the total 17920 retrieved snippets for all claims, 16375 (91.4\%) snippets are from PubMed, with 1297 (7.2\%) from StatPearls and 248 (1.4\%) from Textbooks. Together with the results from the top-k parameter selection experiment in \Cref{sec:topk_selection}, using all three medical corpora is the best option provided by MedRAG to include as many relevant and reliable medical sources as possible for better verification.

\subsection{Baselines' Implementation Details}
\label{sec:baseline_details}

For FActScore decompositions, we prompt \texttt{gpt-4o-mini} with the original prompt from \citet{min-etal-2023-factscore}. 

\citet{song-etal-2024-veriscore} fine-tuned Mistral for claim decomposition, and we use their released codebase and model\footnote{\url{https://huggingface.co/SYX/mistral_based_claim_extractor}} to replicate VeriScore decompositions.

Core filters pre-generated claims, and we evaluate its filtering ability on \methodname\ claims, with the results named as ``\methodname+Core''.
We use the code\footnote{\url{https://github.com/zipJiang/Core}} released by \citet{jiang_core_2024} for Core claim filtering. We used the default settings to remove claims that are not faithful or unique. A claim is considered unfaithful to its original sentence if the sentence is not entailed by the claim, and claims are only unique if they do not entail each other. We do not use the informativeness scorers for additional claim weighting.

\subsection{Claim Decomposition Visualization}
\label{sec:claim_decomp_details}

\begin{figure*}[]
    \centering
    \begin{subfigure}[t]{0.5\textwidth}
        \centering
        \includegraphics[width=\textwidth]{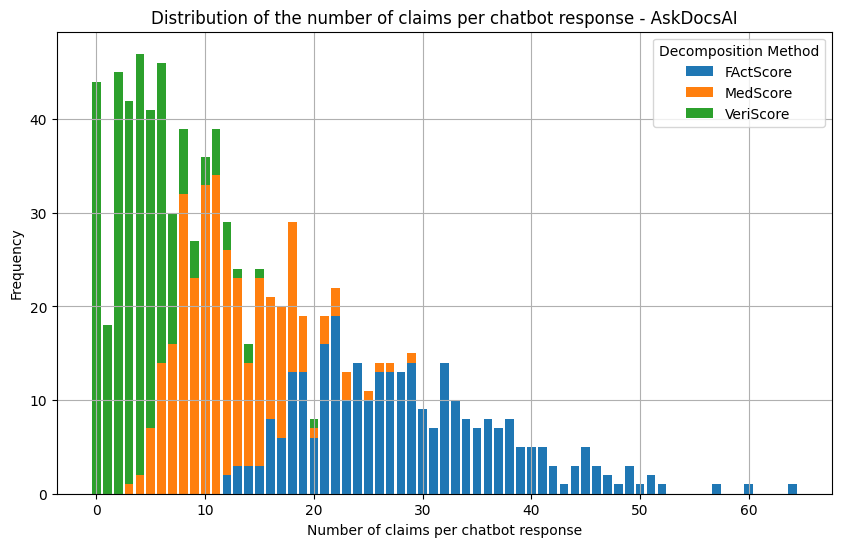}
        \caption{Per chatbot response}
    \end{subfigure}%
    ~
    \begin{subfigure}[t]{0.5\textwidth}
        \centering
        \includegraphics[width=\textwidth]{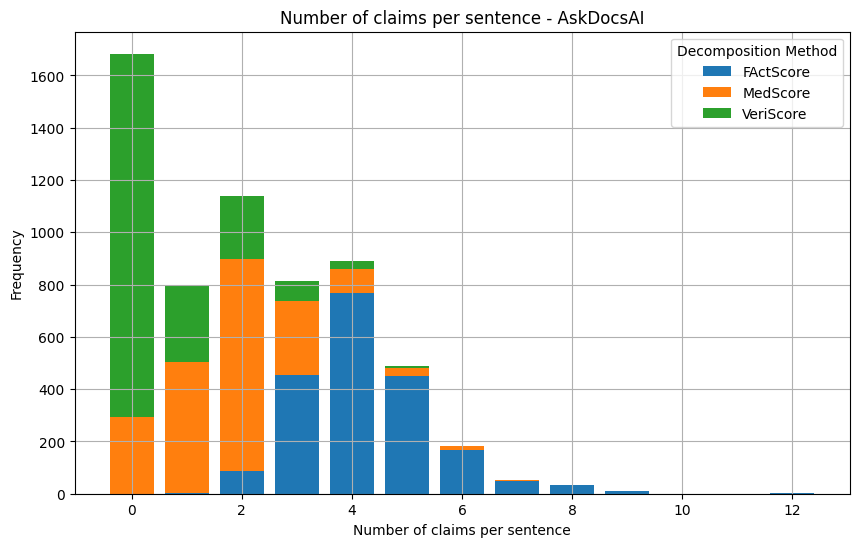}
        \caption{Per sentence}
    \end{subfigure}
    \caption{Number of extracted claims per chatbot response (left) and sentence (right) from FActScore, \methodname, and VeriScore decomposition methods for \datasetname.}
    \label{fig:claim-dist-AskDocs}
\end{figure*}

\paragraph{\datasetname.}The visualization of \Cref{tab:combined_claim_stats} is in \Cref{fig:claim-dist-AskDocs}. \methodname+Core is omitted for figure readability.

\begin{figure*}[]
    \centering
    \begin{subfigure}[t]{0.5\textwidth}
        \centering
        \includegraphics[width=\textwidth]{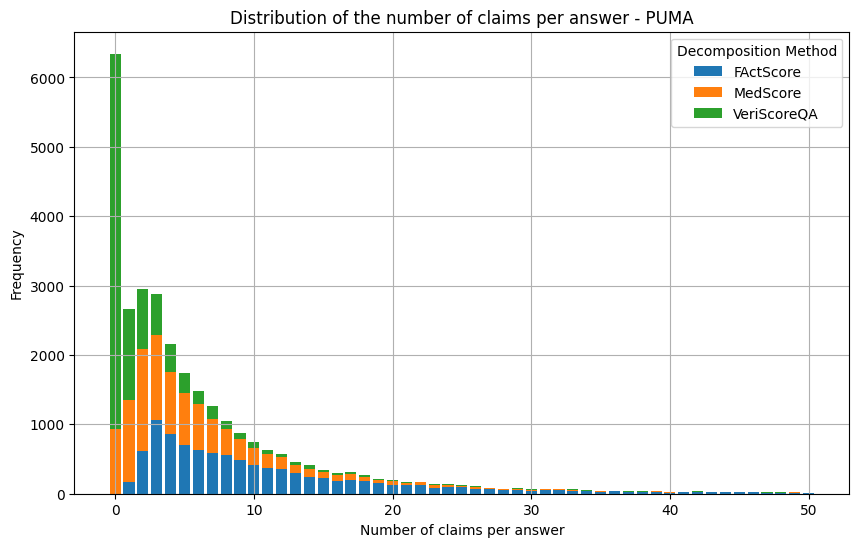}
        \caption{Per answer}
    \end{subfigure}%
    ~
    \begin{subfigure}[t]{0.5\textwidth}
        \centering
        \includegraphics[width=\textwidth]{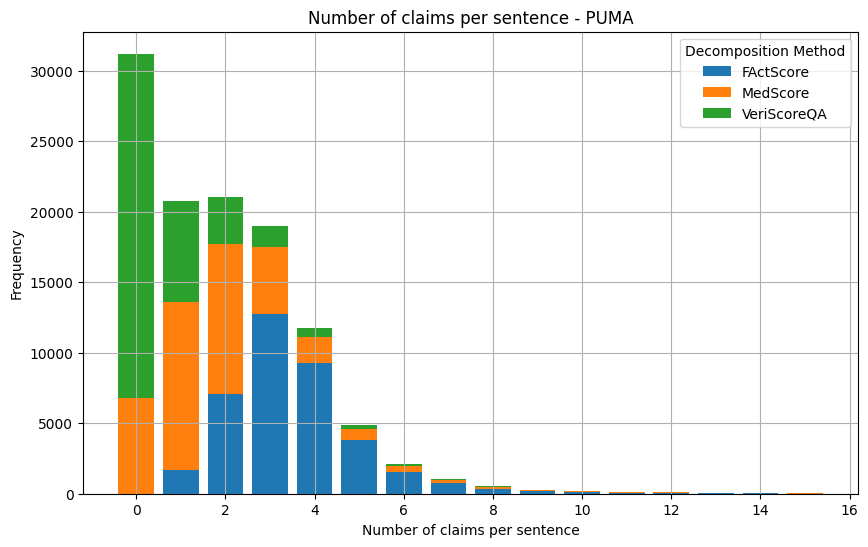}
        \caption{Per sentence}
    \end{subfigure}
    \caption{Number of extracted claims per answer (left) and sentence (right) from FActScore, \methodname, and VeriScoreQA decomposition methods for PUMA.}
    \label{fig:claim-dist-PUMA}
\end{figure*}

\paragraph{PUMA.} The number of claims per answer and number of claims per sentence in the PUMA dataset are clipped for figure readability in \Cref{fig:claim-dist-PUMA}.

\subsection{Verifier Model Selection}
\label{sec:verifier-selection}
There are many available LLMs that could be used as a backbone model for the verification step, so we designed an experiment to evaluate how well common off-the-shelf models can reason about the factuality of a claim given context. In \datasetname, we have a ``ground truth'' in the form of the original doctor's response. Theoretically, any claim decomposed from the doctor's response should be factual when compared to itself, i.e., always True. Therefore, the best verifier model should be able to understand this doctor's response, and only use this response's information to determine whether the claim is True. We compare two open-source LLM families' newly-released models (Llama 3.1-8B, Mistral-7B, and Mistral Small 3) and select the LLM with the highest factuality score as our verifier model. The results are in \Cref{tab:verifier_selection}. The doctor responses are decomposed by MedScore using \texttt{GPT-4o-mini} to ensure the highest claim quality. When verifying doctor responses' claims against the doctor responses, the Mistral Small 3 (\texttt{mistralai/Mistral-Small-24B-Instruct-2501}) model achieves the highest factuality score, 99.35\%.
Previous factuality-related work \citep{min-etal-2023-factscore, wanner_dndscore_2024} mostly use Inst-LLAMA \citep{touvron2023llamaopenefficientfoundation, wang-etal-2022-super} as their verifier model. 

\begin{table}[h]
    \centering
    \begin{tabular}{lc}
    \toprule
        Model & Factuality score \\
    \midrule
        Llama-3.1-8B-Instruct & 97.53\% \\
        Mistral-7B-Instruct-v0.3   &    98.35\% \\
        Mistral Small 3  &  \textbf{99.35\%}\\
    \bottomrule
    \end{tabular}
    \caption{Factuality score by decomposing doctor responses and verifying against doctor responses.}
    \label{tab:verifier_selection}
\end{table}

\subsection{Experiment Computation and Hyperparameters}
\label{sec:compute}
The \methodname\ claim decomposition step with \texttt{gpt-4o-mini} was run serially on one CPU and took 2 hours to decompose all \datasetname\ chatbot responses into claims. The decomposition for both \datasetname\ and PUMA cost \$18USD. 

For the LLM internal knowledge experiments, verification against \texttt{GPT-4o} cost \$10USD. To verify against Llama3-OpenBioLLM-70B, we used four NVIDIA A100 Tensor Core 80GB GPUs. 

External corpus verification with MedCorp required 300GB of RAM and two NVIDIA A100 Tensor Core 80GB GPUs to host Mistral Small 3.

All open-sourced models were hosted with vLLM \citep{kwon2023efficient}.

FActScore and MedScore decompositions are both done by gpt-4o-mini, with temperature=0, top-p sampling=1.0, max token=256. VeriScore decomposition is produced by the author-released model on Huggingface, using the author-released code’s default setting. All verifications use the same hyperparameter setting, with temperature=0, top-p sampling=1.0, max token=256.


\subsection{Annotation Details}
\label{sec:annotation}
The instructions provided to the two author-annotators are below.

The goal is to evaluate the quality of claim decomposition from various models. You will do this by mimicking the LLM decomposition step.

\begin{enumerate}
    \item \textbf{Verifiable or unverifiable?} (From the perspective of Decomposer LLM) If the claim contains any narrative (i.e., patient/doctor experience), or empathic statements (``I’m sorry you’re experiencing pain’’) from the original sentence, then it is unverifiable. Skip to step 3.
    \begin{itemize}
        \item[Ex.] Original sentence: They're looking forward to seeing you at your follow-up appointment and assessing your progress.
        \item[Ex.] Original sentence: Your doctor also wants to remind you that you're just starting your treatment for PFPS, and it's too early to give up hope.
    \end{itemize}

    \item Quality check on verifiable claims (From the perspective of Verifier LLM)
    \begin{enumerate}
        \item \textbf{Wrong structure}. I.e., Imperatives, wrong sentence clause, that makes a verifiable claim hard to be verified against external corpus.
        \begin{itemize}
            \item[Ex.] Imperative: Keep the splint on.
            \item[] Wrong sentence clause: They mentioned that individuals may be told they have low fertility.
            
            \item[Ex.] Imperative: Keep the splint on as directed.
            \item[] Sentence: Instead, focus on following their instructions and keeping the splint on as directed. 
        \end{itemize}

        \item \textbf{Hallucination}. A claim that distorts original sentence meaning, or generates something not mentioned in the sentence. This can also include inferred information not explicitly stated (e.g., “elevating your wrist can help with swelling’’ → “there is swelling in your wrist’’)
        \begin{itemize}
            \item[Ex.] Original sentence: If you have questions or concerns, please let us know.
            \item[] Claim: You may have additional questions or concerns.
            \item[Ex.] Original: Your doctor also wanted to reassure you that even if you do have a scaphoid fracture, most cases can be treated with a cast and do not require surgery.
            \item[] Claim: A scaphoid fracture can occur.
        \end{itemize}

        \item \textbf{Incomplete}. The original sentence has a condition or a modifier AND the claim doesn't have it.
        \begin{itemize}
            \item[Ex.] Original sentence: They mentioned that in many cases, individuals [who are told they have low fertility or trouble getting pregnant] may still be able to conceive.
            \item[] Claim: They mentioned that in many cases, individuals may still be able to conceive.
        \end{itemize}

        \item \textbf{Context dependent}. The sentence contains subject information (e.g., “wrist”, “doctor”) but the claim replaces it with vague pronouns (e.g., “it”, “these”, “they”, “your pain”). Any phrase that is subjective and not fully decontextualized (i.e., ``your specialist’’ versus ``a specialist’’).
        \begin{itemize}
            \item[Ex.] Claim: If you stop doing these exercises, they won't be effective.
            \item[] Original sentence: Your doctor likens these exercises to medicine, meaning that if you stop doing them, they won't be effective in managing your pain.
        \end{itemize}
    \end{enumerate}
    
    \item \textbf{Claim Redundancy} The LLM generates one claim into multiple claims, just by minorly modifying the original correct claim. Mark the simplest claim as redundant.
    \begin{itemize}
        \item[Ex.] Claim 1: They wanted to remind you not to worry too much. (redundant)
        \item[] Claim 2: They wanted to remind you not to worry too much about the day-to-day changes. (redundant too)
        \item[] Claim 3: They wanted to remind you not to worry too much about the day-to-day changes in your symptoms.
    \end{itemize}
\end{enumerate}

\clearpage

\clearpage
\input{tables/AskDocs_prompt}
\input{tables/internal_prompt}
\input{tables/verification_prompt}
\input{tables/NLI_prompt}
\input{tables/MedScore_Balanced_prompt}

\input{tables/ablation_prompt}

\end{document}

%% file: tables/PUMA_NLI_example.tex
\begin{table*}
    \centering
    \begin{tabular}{p{0.13\textwidth} p{0.64\textwidth} p{0.13\textwidth}}
        \toprule
        \multirow{1}{*}{Question} & what is the quickess way to lose weight? &  \\
        \midrule
        
        \multirow{3}{*}{Answer} & not the healthiest but starve yourself or i have heard of using laxatives. however do not recommend. quick weight loss is never healthy & Entailed by verifiable span \\
        \midrule
        \multirow{10}{*}{FActScore}
        & It is not the healthiest option.* \textcolor{red}{\textit{(context-dependent)}} & True \\
        & Starving yourself is mentioned as an option.* \textcolor{red}{\textit{(incomplete)}} & True \\
        & Using laxatives is mentioned as an option.* \textcolor{red}{\textit{(incomplete)}} & True \\
        & The speaker has heard of using laxatives.* \textcolor{red}{\textit{(unverifiable)}} & True \\
        & There is a recommendation being discussed.* \textcolor{red}{\textit{(unverifiable)}} & False \\
        & The recommendation is not being made.* \textcolor{red}{\textit{(unverifiable)}} & True \\
        & The phrase ``however do not recommend'' indicates a contrast or exception to a previous statement.* \textcolor{red}{\textit{(hallucinated)}} & True \\
        & Quick weight loss is not healthy.* \textcolor{red}{\textit{(redundant)}} & True \\
        & Quick weight loss is never healthy. & True \\
        \midrule
        \multirow{4}{*}{\methodname }
        & Starving oneself is not considered a healthy method for weight loss. & True \\
        & Using laxatives is mentioned as a method for weight loss. & True \\
        & Quick weight loss is generally considered unhealthy. & True \\
        \midrule
        VeriScoreQA & No Verifiable Claim *\textcolor{red}{\textit{(omitted)}}& N/A \\
        \bottomrule

    \end{tabular}
    \caption{Examples of claim decomposition and NLI results by FActScore, \methodname\ and VeriScoreQA on PUMA dataset. A claim is labeled as True if it can be inferred from a verifiable answer span, as described in \Cref{sec:PUMA_generalize}. Note that invalid claims marked by * with an invalid category in parentheses can be labeled as True as long as they are entailed by verifiable text spans.}
    \label{tab:PUMA_NLI_example}
\end{table*}

%% file: tables/PUMA_NLI_result.tex
\begin{table*}[]
    \centering
    \begin{tabular}{cccc}
    \toprule 
           &   Verifiable Rate & Adjusted Verifiable Rate$\uparrow$ &Penalization$\downarrow$\\
    \midrule
         FActScore   & 60.09\%&  60.09\% & 0\% \\
         \methodname   & 65.14\% & \textbf{63.75}\% & \textbf{1.4}\%\\
         VeriScoreQA &  82.20\% & 47.48\% & 34.72\%\\
        \bottomrule
    \end{tabular}
    \caption{Automatic claim verifiability evaluation in PUMA with entailment threshold=0.8}
    \label{tab:PUMA_NLI_result}
\end{table*}

%% file: tables/False_Negative_NLI_examples.tex
\begin{table*}[h]
    \centering
    \resizebox{\textwidth}{!}{
    \begin{tabular}{p{0.1\textwidth} p{0.7\textwidth} p{0.13\textwidth}}
        \toprule
        \multirow{1}{{0.1\textwidth}}{Question and Context} & Running a fever - 102.0, to be exact.? What is the best home remedy for a moderate fever?  And - is it feed a cold, starve a fever, or is it the other way around? &  \\
        \midrule
        
        \multirow{3}{*}{Answer1} & Just don't get too hot or cold, cuz I know when you have a fever you feel really hot and then really cold, so don't overheat yourself & Entailed by verifiable span \\
        \midrule
        
        \multirow{3}{*}{\methodname\ }
        & People with a fever may feel really hot. & False \\
        & People with a fever may feel really cold. & False \\
        & It is important for people with a fever to avoid overheating. & False \\
        \midrule
        \multirow{1}{*}{Answer2} & bathe in luke warm water with rubbing alcohol in the water. & \\
        \midrule
        
        \multirow{2}{*}{\methodname }
        & Bathing in lukewarm water is a suggested method for managing a fever. & False \\
        & Rubbing alcohol can be added to lukewarm water. & False \\
        \bottomrule

    \end{tabular}
    }
    \caption{False Negative NLI examples of PUMA claims by \methodname. These claims are valid and verifiable, but labeled as False because the original answers are not correctly labeled with verifiable spans. Both answers here should be labeled with the verifiable span, SUGGESTION.}
    \label{tab:FalseNegative_NLI_example}
\end{table*}

%% file: tables/CalmQA_examples.tex
\begin{table*}[thb]
    \centering
    \resizebox{\textwidth}{!}{
    \begin{tabular}{p{0.15\textwidth} p{0.8\textwidth}}
        \toprule
        \multirow{3}{*}{Question} & Why does the President of the United States need to be born in the United States to be eligible to run? It seems like the country that a person was born in has little to do with their abilities to lead.  \\
        \midrule
        \multirow{4}{0.15\textwidth}{Answer snippet} & It's not a matter of their leadership ability, it's intended to guarantee loyalty. The idea is that most people are loyal to their birth country, so you'd want someone born in the US to be the one acting as commander in chief of the US armed forces, among other things. May not be perfect logic, but it's tradition at this point. $\cdots$ \\
        \midrule
        \multirow{8}{*}{FActScore}
        & It is not a matter of their leadership ability.* \textcolor{red}{\textit{(context-dependent)}}\\
        & It is intended to guarantee loyalty.* \textcolor{red}{\textit{(context-dependent)}} \\
        & Most people are loyal to their birth country. \\
        & The idea is that someone born in the US should act as commander in chief of the US armed forces.* \textcolor{red}{\textit{(context-dependent)}}\\
        & The commander in chief of the US armed forces is an important position.* \textcolor{red}{\textit{(hallucinated)}}\\
        & The US armed forces have a commander in chief.* \textcolor{red}{\textit{(hallucinated)}}\\
        & It may not be perfect logic.* \textcolor{red}{\textit{(context-dependent)}}\\
        & 	It is tradition at this point.* \textcolor{red}{\textit{(context-dependent)}}\\
        \midrule
        \multirow{8}{*}{\methodname }
        & The requirement for the President of the United States to be born in the United States is intended to guarantee loyalty. \\
        &	Most people are loyal to their birth country. \\
        &	It is preferred to have someone born in the US act as commander in chief of the US armed forces.\\
        &	The requirement for the President of the United States to be born in the United States is a tradition.\\
        \midrule
        VeriScoreQA & No Verifiable Claim * \textcolor{red}{\textit{(omitted)}}\\
        \bottomrule

    \end{tabular}
    }
    \caption{Examples of CaLMQA claims by FActScore, \methodname\ and VeriScoreQA. Invalid claims are marked by * with one of their invalid categories in parentheses. One invalid claim can have multiple errors defined in \taxonomy.}
    \label{tab:CaLMQA_example}
\end{table*}

%% file: tables/DnD_example.tex
\begin{table*}[]
    \centering
    \begin{tabular}{p{0.1\textwidth} p{0.85\textwidth}}
        \toprule
        \multirow{2}{0.1\textwidth}{First Sentence} & I spoke to your doctor and they wanted to reassure you that the swelling in your wrist area is a normal part of the healing process.   \\
        \midrule
        
        \multirow{8}{*}{Claims} 
        & \textbf{Subclaim}: I spoke to your doctor.\\
        & \textbf{Decontextualized}: I spoke to the doctor \textit{responsible for your care}.\\
        &\textbf{Subclaim}: Your doctor wanted to reassure you.\\
        &\textbf{Decontextualized}: Your doctor, \textit{who is treating your wrist condition}, wanted to reassure you.\\
        &\textbf{Subclaim}: The swelling in your wrist area is a normal part of the healing process.\\
        &\textbf{Decontextualized}: The swelling in your wrist area, which is part of your healing process \textit{following an injury}, is a normal occurrence.\\
        \toprule
        \multirow{2}{0.1\textwidth}{Second Sentence} & It's expected that the swelling will come and go, and it's not necessarily an indication of the severity of the situation.  \\
        \midrule
        \multirow{10}{*}{Claims} 
        & \textbf{Subclaim}: The swelling in the wrist area is expected to fluctuate.\\
        & \textbf{Decontextualized}: The swelling in the wrist area is expected to come and go \textit{as part of the healing process.}\\
        &\textbf{Subclaim}: The fluctuation of swelling in the wrist area is not necessarily an indication of severity.\\
        &\textbf{Decontextualized}: The fluctuation of swelling in the wrist area is not necessarily an indication of the severity \textit{of the healing situation.}\\
        &\textbf{Subclaim}: The situation being referred to is related to wrist healing.\\
        &\textbf{Decontextualized}: The situation being referred to is related to \textit{the healing process of the wrist injury}.\\

        \toprule
        \multirow{2}{0.1\textwidth}{Third Sentence} & Your doctor emphasized the importance of keeping your wrist elevated, as you've been doing, to help reduce the swelling.\\
        \midrule
        \multirow{15}{*}{Claims} 
        & \textbf{Subclaim}: Your doctor emphasized something.\\
        & \textbf{Decontextualized}: The doctor \textit{who is responsible for your treatment} emphasized something.\\
        &\textbf{Subclaim}: The emphasis was on the importance of keeping your wrist elevated.\\
        &\textbf{Decontextualized}: The emphasis was on the importance of keeping your wrist, \textit{specifically the one that is swollen}, elevated.\\
        &\textbf{Subclaim}: Keeping your wrist elevated is important.\\
        &\textbf{Decontextualized}: Keeping your wrist elevated, \textit{particularly the one with swelling,} is important.\\
        & \textbf{Subclaim}: You have been keeping your wrist elevated.\\
        & \textbf{Decontextualized}: You have been keeping your wrist elevated, \textit{as instructed by your doctor.}\\
        &\textbf{Subclaim}: Keeping your wrist elevated helps reduce swelling.\\
        &\textbf{Decontextualized}: Keeping your wrist elevated helps reduce the swelling \textit{in the wrist area.}\\
        \bottomrule

    \end{tabular}
    \caption{Examples of DnDScore claims from 3 consecutive sentences in a chatbot response in \datasetname. Phrases added for subclaim decontextualization are highlighted in \textit{italics}.}
    \label{tab:DnD_example}
\end{table*}

%% file: tables/AskDocsAI_example.tex
\begin{table*}[]
    \centering
    \begin{tabular}{p{0.15\textwidth} p{0.8\textwidth}}
        \toprule
        \multirow{8}{0.15\textwidth}{User Question} & Post wrist surgery recovery time frame Hi all. Im a 26F, 155lbs, 52, caucasian. I have asthma, allergies, and Mnires disease. Im on Advair, hydrochlorothiazide (for Mnires), Zoloft, and Flonase. I use recreational drugs and alcohol occasionally, no more than once a week. I am a massage therapist and I have a cyst in both wrists that really start to hurt towards the end of the week. I am finally in the position to be able to have them removed (Ive been dealing with this for 5 years in one wrist and 2 years in the other), but I do want to make sure that potential recovery wont be too long. Any estimate would be great! \\
        \midrule
        \multirow{4}{0.15\textwidth}{Doctor Response} & Following a ganglion removal, most surgeons I know recommend a 4-8 week period following surgery prior to returning to high demand physical jobs, such as lifting/pushing/massaging. Check with your surgeon to be sure. They may have specific recommendations based on your case.\\
        \midrule
        \multirow{8}{0.15\textwidth}{Machine Augmented Response}
        & Hi there,

        I spoke to your doctor and they wanted to provide you with some general guidance on the recovery time frame for ganglion removal surgery. Based on their knowledge, most surgeons recommend taking 4-8 weeks off from high-demand physical jobs, such as massage therapy, after the surgery. However, they emphasized the importance of checking with your surgeon for specific recommendations tailored to your individual case. Your surgeon will be able to provide you with more detailed guidance based on your unique situation and the specifics of your surgery.

        Best regards,
        
        [Your Doctor's Name] \\
        \bottomrule

    \end{tabular}
    \caption{The full example of the AskDocsAI dataset shown in \Cref{fig:decompose_verify_overview}}
    \label{tab:AskDocsAI_example}
\end{table*}

%% file: tables/TopK_selection.tex
\begin{table}[h]
    \centering
    {\begin{tabular}{c|c|c}
    \toprule
        Top-k & Factuality Score & Verification Time \\
    \midrule
       Top-5 & 66.88\% & 54.35 min \\
        Top-10 & 70.07\% & 76.22 min\\
        Top-15 & 71.31\% & 86.23 min \\
        Top-20 & 72.23\% & 108.12 min \\
        Top-25 & 72.84\% & 143.63 min \\
       
    \bottomrule
    \end{tabular}}
    \caption{Running time and factuality scores for top-k retrieved medical passages on \datasetname\ with \methodname\ decomposition. The verification time is collected on a single run.}
    \label{tab:topk_selection}
\end{table}

%% file: tables/MedCorp_ablation.tex
\begin{table*}
    \centering
    {\begin{tabular}{c|c|c|c}
    \toprule
        Corpus &Corpus size (\#snippet)& Factuality Score & Verification Time \\
    \midrule
    PubMed-only	& 23.9M &	66.56\% &	59.35 min \\
    StatPearls-only	& 301.2k& 69.34\%&	23.23 min\\
    Textbooks-only &	125.8k &	68.89\% &	25.75 min \\
    StatPearls-and-Textbooks&427k&70.03\%&	25.47 min\\
    All 3 medical corpus	&24.3M	&66.88\%&	54.35 min\\
       
    \bottomrule
    \end{tabular}}
    \caption{Ablation study result of MedCorp individual corpus on \datasetname\ with \methodname\ decomposition. The verification time is collected on a single run.}
    \label{tab:medcorp_ablation_result}
\end{table*}

%% file: tables/AskDocs_prompt.tex
\begin{table*}[]
    \centering
    \begin{tabular}{p{\textwidth}}
    \toprule
    You are a medical assistant acting as an intermediary to a doctor who shares the doctor’s email with the patients.\\
    
    You will be provided with a message written by a patient to the doctor and a draft response written by that doctor. You will provide a written message in response to the patient's question. Your goal is to make the doctor’s response more understandable and digestible for the patient. You are not allowed to come up with your own diagnoses or suggestions for treatment.\\
    
    Your response should use the following guidelines: (a) Write on behalf of the doctor (e.g., “I spoke to your doctor and they recommended”). (b) do not propose any additional clinical content beyond what the doctor has explicitly said (e.g., the doctor writes “try over the counter pain meds” and you enhance it to mention specific types of medications, dosing advice, etc.).\\
    
    This is for a user study of an AI system. We are not asking for medical advice and your response will not be given to a patient. We will only use your response to evaluate AI systems.\\
    \bottomrule
    \end{tabular}
    \caption{Prompt to augment doctor responses with user questions in the AskDocs dataset}
    \label{tab:AskDocs_prompt}
\end{table*}

%% file: tables/internal_prompt.tex
\begin{table*}[t]
    \centering
    \begin{tabular}{p{\linewidth}}
    \toprule
    system prompt: 
    
    You are an assistant who verifies whether a claim from a medical response is True. You should rely exclusively on your own knowledge and always output "True" or "False" first. If there is not enough context or you are unable to verify the claim, then output "False".\\
    \\
    user prompt: 
    
    Using your own knowledge, answer the question.\\
    \\
    Input:\{Your doctor wanted to clarify a few things.\} True or False?\\
    \\
    Output:\\
    \bottomrule
    \end{tabular}
    \caption{Claim verification prompt using LLM internal knowledge. \{claim\} is enclosed in braces.}
    \label{tab:internal-prompt}
\end{table*}

%% file: tables/verification_prompt.tex
\begin{table*}[t]
    \centering
    \begin{tabular}{p{\textwidth}}
    \toprule
    user prompt: 
    
    Answer the question based on the given context.\\
    \\
    \{What tests are you looking for? I think it will depend on what you are asking for. If it's some rare or even uncommon, esoteric study, you're better off seeing a specialist; your PCP may not feel comfortable ordering a test or interpreting it.\}\\
    \\
    Input: \{They may recommend that you see a
    specialist.\} True or False?

    Output: \\ \bottomrule
    \end{tabular}
    \caption{FActScore prompt for verifying extracted facts against a given text. \{Context\} and \{fact\} are enclosed in braces. The context is an original doctor response or 5 retrieved passages from external corpus in our experiments.}
    \label{tab:verification_prompt}
\end{table*}

%% file: tables/NLI_prompt.tex
\begin{table*}[t]
    \centering
    \begin{tabular}{p{\linewidth}}
    \toprule
    system prompt: 
    
    You are an entailment model. You will be given a claim and a premise. You should determine if the premise entails the claim. If the premise entails the claim, return True. If the premise doesn't entail the claim, such as contradiction or neutral, return False. If you are not sure, return False. You should also return the entailment score, which is a float between 0 and 1. The higher the score, the more likely the premise entails the claim. The threshold for the label to be True is the score greater than 0.8. Output in the format of
    
    Label: True/False
    
    Score: 0.6.\\
    \\
    user prompt: 
    
    Does the premise entail this claim?
    \\
    Premise: \{not the healthiest but starve yourself or i have heard of using laxatives. however do not recommend. quick weight loss is never healthy\}\\
    
    Claim: \{Starving oneself is not considered a healthy method for weight loss.\}\\
    \bottomrule
    \end{tabular}
    \caption{NLI prompt for claim verifiability evaluation on PUMA dataset using gpt-4o-mini. \{verifiable span\} and \{claim\} are enclosed in braces.}
    \label{tab:NLI_prompt}
\end{table*}

%% file: tables/MedScore_Balanced_prompt.tex
\onecolumn
\begin{longtable}{p{\linewidth}}
    \toprule
    system prompt:

        You are a medical expert in evaluating how factual a medical sentence is. You break down a sentence into as many facts as possible. The facts should be objective and verifiable against reliable external information such as Wikipedia and PubMed. All subjective personal experiences ("I was or someone did") and personal narratives (stating a past event) are not verifiable and should not be included in the fact list. Facts should be situated within conditions in the sentence. Suggestions (e.g. "I recommend or Your doctor suggest") and opinions (e.g. "I think") should be transformed into objective facts by removing subjective words and pronouns to only retain the core information that can be verified. Imperative instructions ("do something") should be transformed into declarative facts ("doing something is helpful for some conditions").\\
    \\
    If there is an overly specific entity such as "Your partner" or vague references (pronouns, this or that) in the fact, replace it with a general phrase with conditional modifiers using information in the provided context (e.g. "People in some conditions"). Each fact should be verifiable on its own and require no additional context. Do not add additional information outside of the sentence and context. You do not need to justify what you extract.\\
    \\
    If there is no verifiable fact in the sentence, please write "No verifiable claim".\\
    \\
    Here are some examples:\\
    \\
    Context: I spoke to your doctor and they wanted to address your concerns about tetanus. Since you've had your primary tetanus shots as a child, you don't need immunoglobulin (IGG) shots, and they were actually unnecessary during your last visit. $\backslash$n$\backslash$n Considering your tetanus vaccine expired in 2020 and you've got a dirty wound from the Spartan race, your doctor recommends getting a tetanus booster vaccine as soon as possible. They also mentioned that you were due for a booster anyway since it's been more than 3 years since your last vaccine.$\backslash$n$\backslash$nYour doctor is a bit puzzled as to why you were given IGG shots instead of a vaccine during your last visit, but that's not a concern for now. They just want to make sure you get the booster vaccine to be on the safe side. It's best to schedule an appointment for the booster vaccine as soon as possible to avoid any potential risks.\\
    Please breakdown the following sentence into independent facts: I spoke to your doctor and they wanted to address your concerns about tetanus.\\
    Facts:\\
    - No verifiable claim\\
    \\
    Context: I spoke to your doctor, and they expressed concerns about the safety of using anabolic steroids, particularly in combination with the medications your partner is already taking for Addison's disease. The doctor noted that while these substances may have positive effects on muscle and bone health, they also carry significant risks and potential side effects.$\backslash$n$\backslash$nThe doctor mentioned that the anabolic cycle your partner is on is quite intense and requires careful monitoring for potential issues such as infertility, mood swings, and problems related to weight gain, including snoring and possible sleep apnea. They also emphasized the importance of considering the long-term effects of using these substances, particularly when they are stopped.$\backslash$n$\backslash$nThe doctor's primary concern is that your partner's underlying condition, Addison's disease, may not significantly complicate things if well-treated, but it could become an issue when the anabolic cycle is stopped. They strongly advise that your partner consult with a medical professional, ideally their endocrinologist, to discuss the potential risks and consequences of using these substances, especially given their pre-existing condition.$\backslash$n$\backslash$nIt's essential to have an open and honest conversation with a healthcare professional to ensure your partner's safety and well-being. I would encourage you to support your partner in seeking medical advice, and I'm happy to facilitate a discussion with their doctor if needed.\\
    Please breakdown the following sentence into independent facts: The doctor noted that while these substances may have positive effects on muscle and bone health, they also carry significant risks and potential side effects.\\
    Facts:\\
    - Anabolic steroids may have positive effects on muscle health.\\
    - Anabolic steroids may have positive effects on bone health.\\
    - Anabolic steroids may also carry significant risks.\\
    - Anabolic steroids may carry potential side effects.\\
    \\
    Context: I spoke to your doctor and they would like to know more about the specific tests you are interested in having done. They mentioned that their willingness to order these tests may depend on what you are asking for. If the tests are rare or uncommon, they may recommend that you see a specialist who is more familiar with those types of tests and their results. Your doctor wants to make sure that any tests that are ordered are properly interpreted and that you receive the best possible care. Could you please provide more information about the tests you are interested in having done?\\
    Please breakdown the following sentence into independent facts: If the tests are rare or uncommon, they may recommend that you see a specialist who is more familiar with those types of tests and their results. \\
    Facts:\\
    - If the tests are rare or uncommon, seeing a specialist is helpful.\\
    - The specialist is more familiar with rare or uncommon types of tests.\\
    - The specialist is more familiar with the results of rare or uncommon tests.\\
    \\
    Context: Hi there,$\backslash$n$\backslash$nI spoke to your doctor and they wanted to reassure you that getting the immunoglobulin shot within 38 hours of the potential exposure is still considered a safe timeframe. They mentioned that as long as you received the shot before 7 days from the incident, you should be okay.$\backslash$n$\backslash$nYour doctor recommends that you follow up on the scheduled dates for your remaining rabies shots. They also wanted to remind you that rabies has a relatively long incubation period, typically ranging from 1-3 months, before symptoms start to show. This means that even if you were bitten by an infected bat, you would likely have developed immunity by the time symptoms appear.$\backslash$n$\backslash$nPlease try to take care of yourself and manage your anxiety during this time. If you have any further concerns or questions, don't hesitate to reach out.$\backslash$n$\backslash$nBest regards,$\backslash$n$\backslash$n[Your Doctor's Name]\\
    Please breakdown the following sentence into independent facts: They also wanted to remind you that rabies has a relatively long incubation period, typically ranging from 1-3 months, before symptoms start to show.\\
    Facts:\\
    - Rabies has a relatively long incubation period.\\
    - The incubation period for rabies typically ranges from 1-3 months.\\
    - Rabies symptoms start to show after the incubation period.\\
    \\
    Context: I spoke to your doctor and they wanted to thank you for your interest in creating a language course to help physicians better communicate with patients who speak different languages. $\backslash$n$\backslash$nThey mentioned that while language barriers can contribute to the "revolving door syndrome," it's just one of many factors. Other important factors include education, home support, medication noncompliance, and lack of primary care. $\backslash$n$\backslash$nIn terms of a language course, your doctor thinks that Duolingo is a good option. However, they noted that it's challenging for doctors to find the time to learn multiple languages, as there are many languages spoken by patients in their area, including Spanish, Hmong, Chinese, and Polish. They also mentioned that many Spanish-speaking patients have some knowledge of English or have family members who are fluent in English.$\backslash$n$\backslash$nYour doctor didn't specify a preferred medium for the course, but they seemed to appreciate the idea of a convenient and accessible program. They also didn't provide specific vocabulary recommendations, but it's likely that a course focused on medical terminology and common patient interactions would be most useful.\\
    Please breakdown the following sentence into independent facts: However, they noted that it's challenging for doctors to find the time to learn multiple languages, as there are many languages spoken by patients in their area, including Spanish, Hmong, Chinese, and Polish.\\
    Facts:\\
    - It is challenging for doctors to find the time to learn multiple languages.\\
    - Many languages are spoken by patients in doctors' area.\\
    - Spanish is one of the languages spoken by patients in doctors' area.\\
    - Hmong is one of the languages spoken by patients in doctors' area.\\
    - Chinese is one of the languages spoken by patients in doctors' area.\\
    - Polish is one of the languages spoken by patients in doctors' area.\\
    \\
    Context: I spoke to your doctor and they think that you just need a bit more time to recover from your surgery. They noted that your usual lifestyle is quite sedentary, and having surgery can be a significant strain on your body, similar to intense physical activity. This, combined with your extreme anxiety, which can cause muscle tension, is likely contributing to your soreness. $\backslash$n$\backslash$nAs long as you don't develop a fever and your wounds show no signs of infection, your doctor believes that there's not much more the hospital can do for you that you can't do at home. Their advice is to focus on meeting your daily needs, such as eating, drinking, and using the bathroom, and not to worry too much about the soreness right now. $\backslash$n$\backslash$nOnce the soreness starts to subside, they recommend that you try to gradually increase your activities, starting with small steps like sitting in a chair, standing, and eventually walking, until you're back to your normal self.\\
    Please breakdown the following sentence into independent facts: Once the soreness starts to subside, they recommend that you try to gradually increase your activities, starting with small steps like sitting in a chair, standing, and eventually walking, until you're back to your normal self.\\
    Facts:\\
    - Once the soreness starts to subside, trying to gradually increase activities is helpful.\\
    - People with soreness should start with small steps.\\
    - Small steps for people with soreness include sitting in a chair.\\
    - Small steps for people with soreness include standing.\\
    - Small steps for people with soreness include walking.\\
    - Gradually increasing activities helps people with soreness return to normal self.\\
    \\
    Context: I spoke to your doctor and they wanted to address your concerns regarding the leakage you experienced after your bowel surgery in 2013. According to them, it is possible for an abnormal connection to form between your bowel and your bladder or vagina, which is known as a fistula. This could potentially cause the leakage of substances from your bowel into your urinary tract or vagina.$\backslash$n$\backslash$nYour doctor recommends reviewing the notes from your second surgery to understand the nature of the repairs that were performed. This information may help clarify what happened in your specific case.$\backslash$n$\backslash$nRegarding your concerns about the quality of care you received from your initial surgeon, your doctor advises that medical malpractice is a complex issue that depends on many factors, including the specific circumstances of your case and the laws in your location. If you're interested in exploring this further, they recommend consulting with a lawyer who can provide guidance on whether you have a valid case.$\backslash$n$\backslash$nPlease let us know if you have any further questions or concerns, and we'll be happy to help.\\
    Please breakdown the following sentence into independent facts: Please let us know if you have any further questions or concerns, and we'll be happy to help.\\
    Facts:\\
    - No verifiable claim\\
    \\
    Context: Dear Patient,$\backslash$n$\backslash$nI spoke to your doctor and they wanted to address your concerns about the sharp pain you've been experiencing above your left eyebrow when exposed to strong wind. After reviewing your symptoms, they think that trigeminal neuralgia could be a possible cause of your pain. $\backslash$n$\backslash$nThey considered cluster headaches as well, but they believe it's less likely in your case because cluster headaches typically occur at the same time every day, last for several days in a row, and involve pain behind or around the entire eye on one side.$\backslash$n$\backslash$nYour doctor would like to discuss this further with you and explore the possibility of trigeminal neuralgia being the cause of your pain. They recommend that you schedule a follow-up appointment to discuss your symptoms in more detail and determine the best course of action.$\backslash$n$\backslash$nPlease let us know if you have any questions or concerns before your appointment.$\backslash$n$\backslash$nBest regards,$\backslash$n[Your Name]$\backslash$nOn behalf of [Doctor's Name]\\
    Please breakdown the following sentence into independent facts: Dear Patient,$\backslash$n$\backslash$nI spoke to your doctor and they wanted to address your concerns about the sharp pain you've been experiencing above your left eyebrow when exposed to strong wind.\\
    Facts:\\
    - No verifiable claim\\
    \\
    Context: I spoke to your doctor, and they recommended that you visit a pharmacy to get an over-the-counter anti-nausea medication, such as Dramamine (also known as Gravol), to help alleviate your symptoms. They would like to know how you're feeling now, 4 hours after your initial message, to assess if your condition is improving or if further action is needed. If your symptoms persist, your doctor may want to investigate further to determine the cause of your discomfort. Please let us know your current status so we can provide further guidance.\\
    Please breakdown the following sentence into independent facts: They would like to know how you're feeling now, 4 hours after your initial message, to assess if your condition is improving or if further action is needed.\\
    Facts:\\
    - No verifiable claim\\
    \\
    Context: I spoke to your doctor and they wanted to reassure you that, given you are not sexually active and have never had penetrative intercourse, it is not possible for you to be pregnant. They understand that you have been experiencing anxiety and pregnancy scares, and they think it's a great idea for you to see a psychologist to help you manage these feelings.$\backslash$n$\backslash$nRegarding your symptoms, your doctor believes that starting a stable contraceptive therapy, such as the pill or other hormonal methods, could be helpful in regulating your periods and alleviating some of the symptoms you're experiencing. They think this could be a useful approach to help you feel better and more in control of your situation.$\backslash$n$\backslash$nPlease keep in mind that your upcoming ultrasound will likely provide more insight into what's going on with your body, and your doctor will be able to discuss the results with you and determine the best course of action.\\
    Please breakdown the following sentence into independent facts: They understand that you have been experiencing anxiety and pregnancy scares, and they think it's a great idea for you to see a psychologist to help you manage these feelings.\\
    Facts:\\
    - Seeing a psychologist can help manage feelings of anxiety and pregnancy scares.\\
    \\
    user prompt:
    
    Context:\{\}\\
    Please breakdown the following sentence into independent facts:\{\}\\
    Facts:\\
    \bottomrule
    \\
    \caption{Instructions and 10 balanced ICL examples in the decomposition prompt of \methodname.}
    \label{tab:MedScorePrompt}
\end{longtable}
\twocolumn

%% file: tables/ablation_prompt.tex
\onecolumn
\begin{longtable}{p{\linewidth}}
    \toprule
    \\
    Context: I spoke to your doctor and they wanted to address your concerns about tetanus. Since you've had your primary tetanus shots as a child, you don't need immunoglobulin (IGG) shots, and they were actually unnecessary during your last visit. $\backslash$n$\backslash$n Considering your tetanus vaccine expired in 2020 and you've got a dirty wound from the Spartan race, your doctor recommends getting a tetanus booster vaccine as soon as possible. They also mentioned that you were due for a booster anyway since it's been more than 3 years since your last vaccine.$\backslash$n$\backslash$nYour doctor is a bit puzzled as to why you were given IGG shots instead of a vaccine during your last visit, but that's not a concern for now. They just want to make sure you get the booster vaccine to be on the safe side. It's best to schedule an appointment for the booster vaccine as soon as possible to avoid any potential risks.\\
    Please breakdown the following sentence into independent facts: I spoke to your doctor and they wanted to address your concerns about tetanus.\\
    Facts:\\
    - No verifiable claim\\
    \\
    Context: I spoke to your doctor, and they expressed concerns about the safety of using anabolic steroids, particularly in combination with the medications your partner is already taking for Addison's disease. The doctor noted that while these substances may have positive effects on muscle and bone health, they also carry significant risks and potential side effects.$\backslash$n$\backslash$nThe doctor mentioned that the anabolic cycle your partner is on is quite intense and requires careful monitoring for potential issues such as infertility, mood swings, and problems related to weight gain, including snoring and possible sleep apnea. They also emphasized the importance of considering the long-term effects of using these substances, particularly when they are stopped.$\backslash$n$\backslash$nThe doctor's primary concern is that your partner's underlying condition, Addison's disease, may not significantly complicate things if well-treated, but it could become an issue when the anabolic cycle is stopped. They strongly advise that your partner consult with a medical professional, ideally their endocrinologist, to discuss the potential risks and consequences of using these substances, especially given their pre-existing condition.$\backslash$n$\backslash$nIt's essential to have an open and honest conversation with a healthcare professional to ensure your partner's safety and well-being. I would encourage you to support your partner in seeking medical advice, and I'm happy to facilitate a discussion with their doctor if needed.\\
    Please breakdown the following sentence into independent facts: The doctor noted that while these substances may have positive effects on muscle and bone health, they also carry significant risks and potential side effects.\\
    Facts:\\
    - Anabolic steroids may have positive effects on muscle health.\\
    - Anabolic steroids may have positive effects on bone health.\\
    - Anabolic steroids may also carry significant risks.\\
    - Anabolic steroids may carry potential side effects.\\
    \\
    Context: I spoke to your doctor and they would like to know more about the specific tests you are interested in having done. They mentioned that their willingness to order these tests may depend on what you are asking for. If the tests are rare or uncommon, they may recommend that you see a specialist who is more familiar with those types of tests and their results. Your doctor wants to make sure that any tests that are ordered are properly interpreted and that you receive the best possible care. Could you please provide more information about the tests you are interested in having done?\\
    Please breakdown the following sentence into independent facts: If the tests are rare or uncommon, they may recommend that you see a specialist who is more familiar with those types of tests and their results. \\
    Facts:\\
    - If the tests are rare or uncommon, seeing a specialist is helpful.\\
    - The specialist is more familiar with rare or uncommon types of tests.\\
    - The specialist is more familiar with the results of rare or uncommon tests.\\
    \\
    Context: Hi there,$\backslash$n$\backslash$nI spoke to your doctor and they wanted to reassure you that getting the immunoglobulin shot within 38 hours of the potential exposure is still considered a safe timeframe. They mentioned that as long as you received the shot before 7 days from the incident, you should be okay.$\backslash$n$\backslash$nYour doctor recommends that you follow up on the scheduled dates for your remaining rabies shots. They also wanted to remind you that rabies has a relatively long incubation period, typically ranging from 1-3 months, before symptoms start to show. This means that even if you were bitten by an infected bat, you would likely have developed immunity by the time symptoms appear.$\backslash$n$\backslash$nPlease try to take care of yourself and manage your anxiety during this time. If you have any further concerns or questions, don't hesitate to reach out.$\backslash$n$\backslash$nBest regards,$\backslash$n$\backslash$n[Your Doctor's Name]\\
    Please breakdown the following sentence into independent facts: They also wanted to remind you that rabies has a relatively long incubation period, typically ranging from 1-3 months, before symptoms start to show.\\
    Facts:\\
    - Rabies has a relatively long incubation period.\\
    - The incubation period for rabies typically ranges from 1-3 months.\\
    - Rabies symptoms start to show after the incubation period.\\
    \\
    Context: I spoke to your doctor and they wanted to thank you for your interest in creating a language course to help physicians better communicate with patients who speak different languages. $\backslash$n$\backslash$nThey mentioned that while language barriers can contribute to the "revolving door syndrome," it's just one of many factors. Other important factors include education, home support, medication noncompliance, and lack of primary care. $\backslash$n$\backslash$nIn terms of a language course, your doctor thinks that Duolingo is a good option. However, they noted that it's challenging for doctors to find the time to learn multiple languages, as there are many languages spoken by patients in their area, including Spanish, Hmong, Chinese, and Polish. They also mentioned that many Spanish-speaking patients have some knowledge of English or have family members who are fluent in English.$\backslash$n$\backslash$nYour doctor didn't specify a preferred medium for the course, but they seemed to appreciate the idea of a convenient and accessible program. They also didn't provide specific vocabulary recommendations, but it's likely that a course focused on medical terminology and common patient interactions would be most useful.\\
    Please breakdown the following sentence into independent facts: However, they noted that it's challenging for doctors to find the time to learn multiple languages, as there are many languages spoken by patients in their area, including Spanish, Hmong, Chinese, and Polish.\\
    Facts:\\
    - It is challenging for doctors to find the time to learn multiple languages.\\
    - Many languages are spoken by patients in doctors' area.\\
    - Spanish is one of the languages spoken by patients in doctors' area.\\
    - Hmong is one of the languages spoken by patients in doctors' area.\\
    - Chinese is one of the languages spoken by patients in doctors' area.\\
    - Polish is one of the languages spoken by patients in doctors' area.\\
    \\
    Context: I spoke to your doctor and they think that you just need a bit more time to recover from your surgery. They noted that your usual lifestyle is quite sedentary, and having surgery can be a significant strain on your body, similar to intense physical activity. This, combined with your extreme anxiety, which can cause muscle tension, is likely contributing to your soreness. $\backslash$n$\backslash$nAs long as you don't develop a fever and your wounds show no signs of infection, your doctor believes that there's not much more the hospital can do for you that you can't do at home. Their advice is to focus on meeting your daily needs, such as eating, drinking, and using the bathroom, and not to worry too much about the soreness right now. $\backslash$n$\backslash$nOnce the soreness starts to subside, they recommend that you try to gradually increase your activities, starting with small steps like sitting in a chair, standing, and eventually walking, until you're back to your normal self.\\
    Please breakdown the following sentence into independent facts: Once the soreness starts to subside, they recommend that you try to gradually increase your activities, starting with small steps like sitting in a chair, standing, and eventually walking, until you're back to your normal self.\\
    Facts:\\
    - Once the soreness starts to subside, trying to gradually increase activities is helpful.\\
    - People with soreness should start with small steps.\\
    - Small steps for people with soreness include sitting in a chair.\\
    - Small steps for people with soreness include standing.\\
    - Small steps for people with soreness include walking.\\
    - Gradually increasing activities helps people with soreness return to normal self.\\
    \\
    Context: I spoke to your doctor and they wanted to address the questions you have regarding your loved one's complications from COVID-19. The doctor believes that the likely sequence of events is that the COVID-19 infection led to demand ischemia, which in turn caused the myocardial infarction (MI). $\backslash$n$\backslash$nThe doctor thinks that both the transfer hospital and the receiving hospital properly prioritized the patient's issues and treated the most life-threatening condition, the acute MI, first. Unfortunately, the patient had many underlying risk factors that made them more susceptible to severe illness from any infection, not just COVID-19.$\backslash$n$\backslash$nRegarding the patient going without clopidogrel for 10 days, the doctor agrees that this may have contributed to the MI, although it's impossible to determine the exact extent of its impact.$\backslash$n$\backslash$nAs for the new diagnoses of congestive heart failure (CHF), chronic obstructive pulmonary disease (COPD), and acute respiratory failure, the doctor did not provide a specific prognosis. However, they did mention that the patient's underlying health conditions and the severity of their illness have made their situation more challenging.$\backslash$n$\backslash$nThe doctor also believes that initiating COVID-19 treatment at the time of presentation may not have significantly altered the course of the patient's illness.$\backslash$n$\backslash$nPlease let us know if you have any further questions or concerns.\\
    Please breakdown the following sentence into independent facts: However, they did mention that the patient's underlying health conditions and the severity of their illness have made their situation more challenging. \\
    Facts:\\
    - The patient's underlying health conditions makes the acute myocardial infarction situation more challenging.\\
    - The severity of the patient's illness makes the acute myocardial infarction situation more challenging.\\
    - The patient's acute myocardial infarction situation is challenging.\\
      \\
    Context: I spoke to your doctor and they wanted to address your concerns regarding the leakage you experienced after your bowel surgery in 2013. According to them, it is possible for an abnormal connection to form between your bowel and your bladder or vagina, which is known as a fistula. This could potentially cause the leakage of substances from your bowel into your urinary tract or vagina.$\backslash$n$\backslash$nYour doctor recommends reviewing the notes from your second surgery to understand the nature of the repairs that were performed. This information may help clarify what happened in your specific case.$\backslash$n$\backslash$nRegarding your concerns about the quality of care you received from your initial surgeon, your doctor advises that medical malpractice is a complex issue that depends on many factors, including the specific circumstances of your case and the laws in your location. If you're interested in exploring this further, they recommend consulting with a lawyer who can provide guidance on whether you have a valid case.$\backslash$n$\backslash$nPlease let us know if you have any further questions or concerns, and we'll be happy to help.\\
    Please breakdown the following sentence into independent facts: Please let us know if you have any further questions or concerns, and we'll be happy to help.\\
    Facts:\\
    - No verifiable claim\\
    \\

    user prompt:
    
    Context:\{\}\\
    Please breakdown the following sentence into independent facts:\{\}\\
    Facts:\\
    \bottomrule
    \\
    \caption{8 ICL examples used in the ablation study of \methodname\ decomposition prompt.}
    \label{tab:MedScore_Ablation_Prompt}

\end{longtable}
\twocolumn